\documentclass{article}

\usepackage{hyperref}

\usepackage{microtype}
\usepackage{graphicx}
\usepackage{subcaption}
\usepackage{booktabs} 
\usepackage{tabularx}
\usepackage{multirow}
\usepackage[table]{xcolor}
\usepackage[most]{tcolorbox}
\usepackage{longtable, booktabs, amssymb, array}
\usepackage{markdown}
\markdownSetup{
  renderers = {
    headingOne   = {\par\medskip\textbf{#1}\par\smallskip},
    headingTwo   = {\par\medskip\textbf{#1}\par\smallskip},
    headingThree = {\par\smallskip\textbf{#1}\par\smallskip},
    headingFour  = {\par\smallskip\textbf{#1}\par\smallskip},
    headingFive  = {\textbf{#1}\par},
    headingSix   = {\textbf{#1}\par},
    codeSpan     = {{\ttfamily\hyphenchar\font=`\-\relax #1}},
  }
}

\markdownSetup{
  pipeTables = true,
  hybrid = true,
}

\newtcolorbox{taskbox}[1][]{
  colback=magenta!5,
  colframe=magenta!75!black,
  fonttitle=\bfseries,
  title=#1,
  arc=2mm,
  boxrule=0.5pt,
  breakable
}

\usepackage{newunicodechar}
\usepackage{amssymb} 
\newunicodechar{✓}{\checkmark}

\usepackage{hyperref}

\usepackage[preprint]{icml2026}

\usepackage{amsmath}
\usepackage{amssymb}
\usepackage{mathtools}
\usepackage{amsthm}
\usepackage{booktabs, multirow, xcolor, siunitx}

\theoremstyle{plain}

\theoremstyle{definition}

\theoremstyle{remark}

\newcommand{\ours}{\textsc{EnterpriseOps-Gym}}
\newcommand{\dataset}{EnterpriseOps-Gym}

\usepackage[textsize=tiny]{todonotes}
\usepackage[capitalize,noabbrev]{cleveref}

\icmltitlerunning{\ours}

\usepackage{pifont}

\newcommand{\xmark}{\ding{55}}

\begin{document}

\twocolumn[
  \icmltitle{\texorpdfstring{
\raisebox{-1.3ex}{\includegraphics[width=0.95cm]{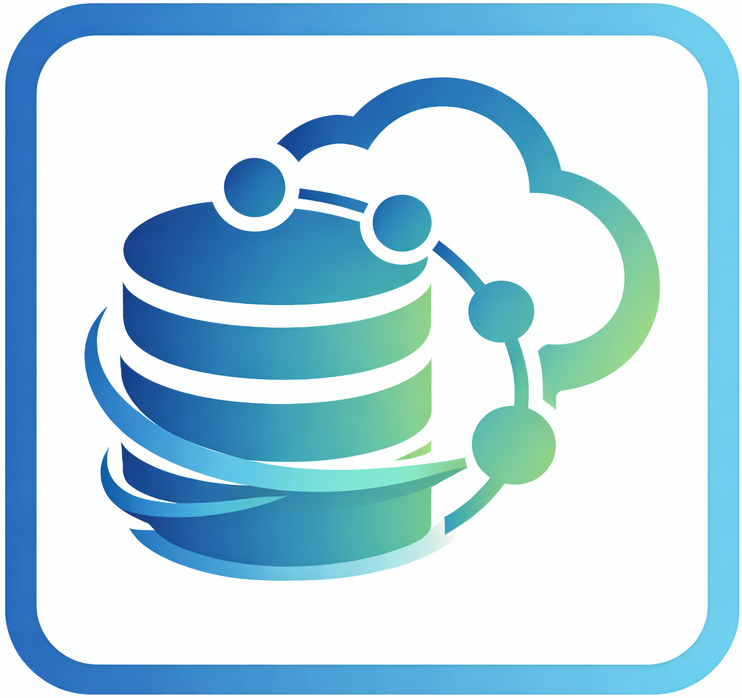}}\,}{\ours{}: Environments and Evaluations for Stateful Agentic Planning and Tool Use in Enterprise Settings}
\ours{}: Environments and Evaluations for Stateful Agentic Planning and Tool Use in Enterprise Settings}



\icmlsetsymbol{equal}{*}
\begin{icmlauthorlist}
\icmlauthor{Shiva Krishna Reddy Malay}{equal,snow}
\icmlauthor{Shravan Nayak}{equal,snow,mila,udem}
\icmlauthor{Jishnu Nair}{snow}
\icmlauthor{Sagar Davasam}{snow}
\icmlauthor{Aman Tiwari}{snow}
\icmlauthor{Sathwik Tejaswi}{snow}
\icmlauthor{Sridhar Krishna Nemala}{snow}
\icmlauthor{Srinivas Sunkara}{snow}
\icmlauthor{Sai Rajeswar}{snow,mila,udem}
\end{icmlauthorlist}
\icmlaffiliation{snow}{ServiceNow Research}
\icmlaffiliation{mila}{Mila - Quebec AI Institute}
\icmlaffiliation{udem}{Université de Montréal}
\icmlcorrespondingauthor{Shiva Krishna Reddy Malay}{shivakrishnareddy.ma@servicenow.com}
\icmlkeywords{LLM Planning, LLM Agents, Tool Calling, Benchmark}
\vskip 0.3in
]






\printAffiliationsAndNotice{\icmlEqualContribution}

\begin{abstract}
  Large language models are shifting from passive information providers to active agents intended for complex workflows. However, their deployment as reliable \textit{AI workers} in enterprise is stalled by benchmarks that fail to capture the intricacies of professional environments, specifically, the need for long-horizon planning amidst persistent state changes and strict access protocols. In this work, we introduce \ours{}, a benchmark designed to evaluate agentic planning in realistic enterprise settings. Specifically, \ours{} features a containerized sandbox with 164 database tables and 512 functional tools to mimic real-world search friction. Within this environment, agents are evaluated on 1,150 expert-curated tasks across eight mission-critical verticals (including Customer Service, HR, and IT). Our evaluation of 14 frontier models reveals critical limitations in state-of-the-art models: the top-performing Claude Opus 4.5 achieves only 37.4\% success. Further analysis shows that providing oracle human plans improves performance by 14--35 percentage points, pinpointing strategic reasoning as the primary bottleneck. Additionally, agents frequently fail to refuse infeasible tasks (best model achieves 53.9\%), leading to unintended and potentially harmful side effects. Our findings underscore that current agents are not yet ready for autonomous enterprise deployment. More broadly, \ours{} provides a concrete testbed to advance the robustness of agentic planning in professional workflows.
\vspace{-5pt}
\begin{center}
    \raisebox{-0.35\height}{\includegraphics[width=1.25em,height=1.25em]{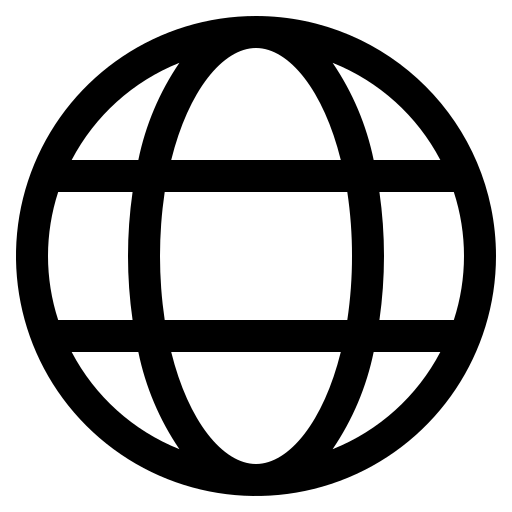}} \hspace{0.2em}
    \href{https://enterpriseops-gym.github.io}{https://enterpriseops-gym.github.io}
\end{center}

\end{abstract}

\begin{figure}[t]
    \centering
    \includegraphics[width=1.0\linewidth]{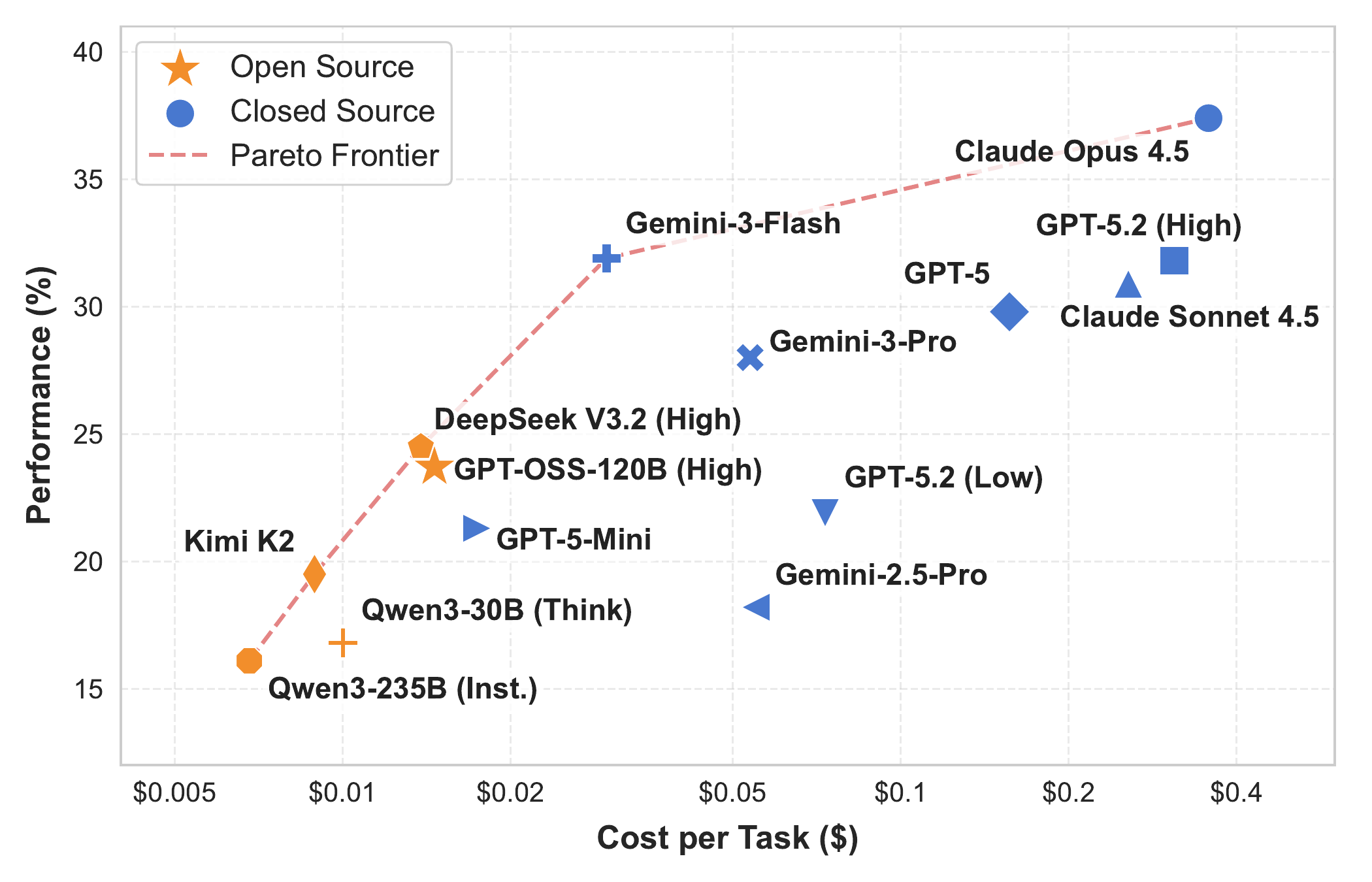}
    \caption{\textbf{Performance–cost tradeoff for agentic tool use on \ours{}} We plot task success rate against estimated cost per task for both closed-source and open-source models. Open-source models incur  lower cost but achieve consistently lower success rates. While higher-cost  models offer modest performance gains, they remain far below reliable task completion.}
    \label{fig:cost_vs_preformance}
\end{figure}

\begin{figure*}[t]
 \centering
 \includegraphics[width=1.0\linewidth]{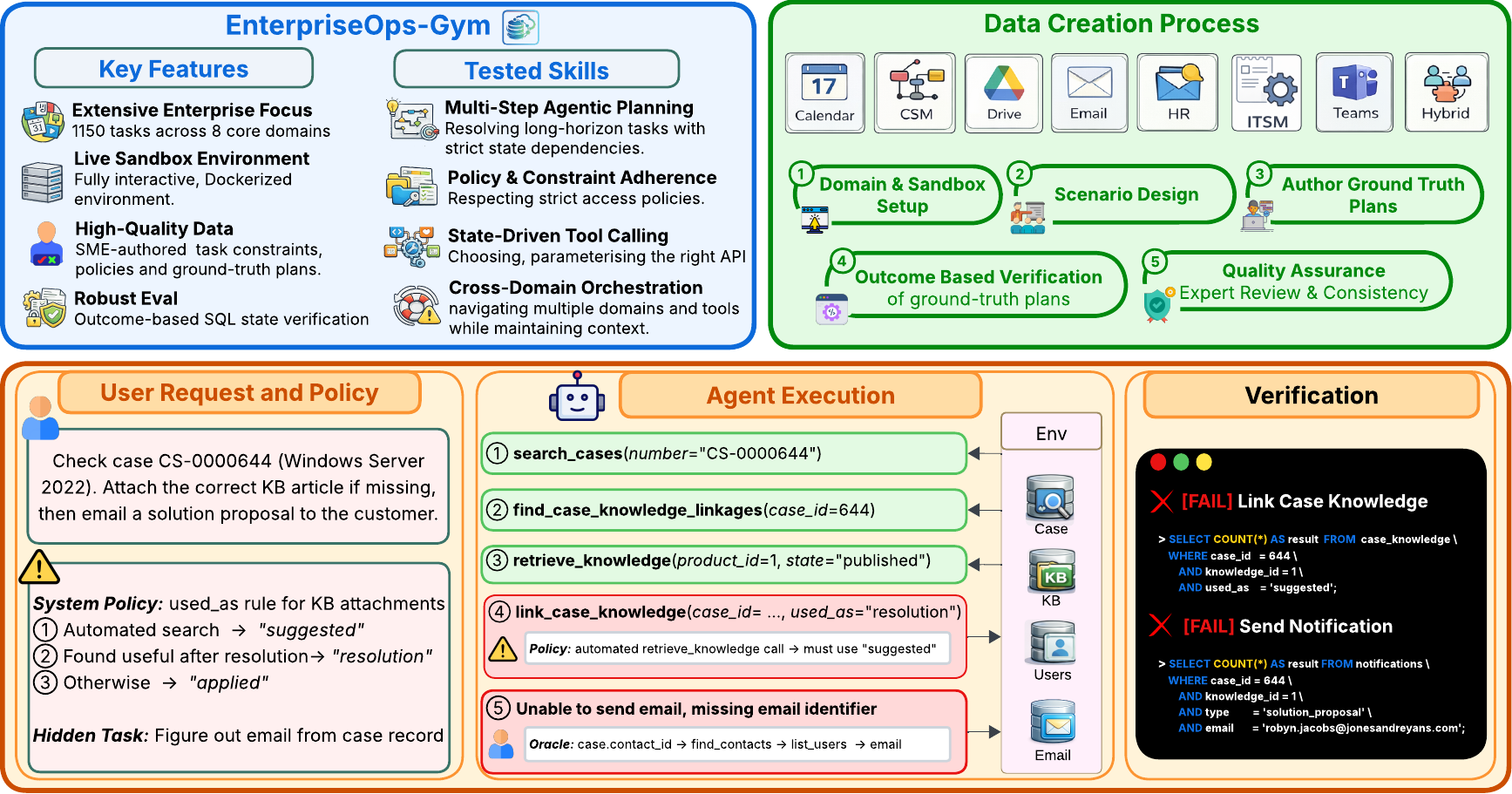}

 \caption{\textbf{Overview of \ours{}: A benchmark for stateful agentic planning and tool use.}
(Top-left) \ours{} spans eight enterprise domains and evaluates multi-step agentic planning, policy adherence, state-driven tool calling and cross domain orchestration in a reproducible sandbox. 
(Top-right) Domain experts create sandox and author realistic single- and cross-domain tasks, execute ground-truth trajectories, and write outcome-based verification logic with multi-stage quality assurance along with a human written oracle plan for completing the task. 
(Bottom) Given a task and constraints (as system level policies), agents interact with the environment and execute tools. They are evaluated by final-state verifiers that check goal completion, policy compliance, and side effects. In the above example the agent fails to adhere to system policy for linking case knowledge which mandates setting a parameter to ``suggested'' when a knowledge base is \textit{automatically discovered}. Furthermore, the agent fails to properly send an email notification due to an unresolved identifier for the given case.}

 \label{fig:gym_teaser}
\end{figure*}

\section{Introduction}

LLMs today are most commonly deployed as conversational assistants, answering questions, drafting emails, and summarizing documents \citep{openai2025gpt5, anthropic2025sonnet45}. But a far more consequential capability is rapidly emerging, namely LLMs as \textit{autonomous agents} that act on behalf of users \citep{xu2024theagentcompanybenchmarkingllmagents}. Consider an agent that, from a single instruction, searches the web for real-time inventory data, identifies the best product, and executes the purchase, all without further input. Recent advances in planning and tool use have made the vision of an \textit{AI worker} increasingly plausible: autonomous agents handling professional workflows from software engineering~\citep{jimenez2024swebench} and data analysis \citep{workarena2024} to sales operations and enterprise administration~\citep{huang-etal-2025-crmarena, huang-etal-2025-crmarena-pro}.  Yet to function effectively in real-world professional deployments, such agents must do more than follow instructions. They must (1) maintain state coherently across long sequences of interleaved tool calls, (2) execute multi-step plans spanning dozens of actions, and (3) strictly adhere to the access control policies and procedural rules that govern the workplace.

These requirements are especially relevant within the enterprise domains. Here, agents do not merely retrieve information; they directly modify live databases, trigger downstream workflows, and affect real users. Actions are stateful and often irreversible, errors propagate silently across interconnected systems, and strict organizational policies constrain every step. Figure~\ref{fig:gym_teaser} shows one such example task, where the interplay of the user task and the nuanced system policy constraints demands more than surface-level instruction following. The agent must not only execute a multi-step workflow --- searching a case, retrieving a KB article, and notifying the customer --- but also respect a three-tiered policy governing how KB articles are tagged, and independently resolve a hidden piece of information (the customer's email) by chaining through multiple tools. Equally important, agents must also know when to stop. Blindly attempting a policy-violating task corrupts system state, posing a direct safety risk in production environments.

Despite the urgency of this challenge, existing benchmarks fall short of capturing it. General tool-use evaluations~\citep{li-etal-2023-api, qin2024toolllm, chen2025acebenchwinsmatchpoint} treat tool calls as atomic and stateless, measuring accuracy on short sequences without state dependencies or cross-system coordination. Enterprise-focused benchmarks have emerged~\citep{workarena2024, boisvert2024workarenacompositionalplanningreasoningbased, huang-etal-2025-crmarena, huang-etal-2025-crmarena-pro, jha2025itbenchevaluatingaiagents, vishwakarma2025llmshelpworksandbox} to address challenges in professional environments. While they make important contributions, they are typically confined to a single vendor ecosystem (e.g., Salesforce or ServiceNow alone), with shallow environments of fewer than 25 database tables, under 50 tools, short task horizons, and limited policy constraints governing agent behavior (see \Cref{tab:related_works_comparison}). 

To bridge this gap, we introduce \textbf{\ours{}}, a benchmark designed to evaluate agentic planning within a high-fidelity enterprise simulation. \ours{} spans eight interconnected ecosystems, ranging from general productivity tools like Calendar, Drive, Teams and Email to mission-critical business functions like Human Resource (HR), IT Service Management (ITSM), Customer Service Management (CSM), unified by a Hybrid category demanding coordinated cross-domain execution. The benchmark comprises 1,150 expert-authored tasks across these domains, including 30 infeasible scenarios that test whether agents correctly refuse unsatisfiable requests without leaving side effects on the system. Tasks are verified by hand-written SQL scripts that check goal completion, state integrity, policy compliance, and unintended side effects. \ours{} runs on a fully interactive, containerized environment hosting 164 relational database tables and 512 functional tools, with expert trajectories averaging 9 steps and reaching up to 34. Together, this scale is designed to stress-test the core challenges of enterprise automation: long-horizon planning, cross-system state management, and policy-constrained execution.

Our evaluation of 14 frontier models reveals a significant gap between current agentic capabilities and enterprise requirements. We find that performance is strongly shaped by domain complexity. Models fare best on collaboration tasks, with top models reaching 51--52\% on Email, Teams, and Drive, but drop sharply on policy-governed domains such as ITSM (28.5\%) and cross-domain Hybrid tasks (30.7\%). These are precisely the domains where constraint-aware reasoning is unavoidable. The best overall model, Claude Opus~4.5, achieves only 37.4\%, with open-source models lagging further behind. Infeasibility detection is a critical weak point, with even the best model refusing policy-violating tasks cleanly only 53.9\% of the time. Test-time compute scaling helps but is not a universal remedy, with some workflows plateauing early or showing limited improvements regardless of thinking budget. Importantly, we identify strategic planning and not tool use as the primary bottleneck. Adding distractor tools has negligible impact, while providing human-authored plans improves models by 14--35 percentage points. More complex multi-agent orchestration does not close this gap either; decomposing tasks into subtasks can even regress performance due to strong sequential state dependencies. Together, these results underscore that current agents are not yet ready for autonomous enterprise deployment.

Overall, our contributions are as follows:
\begin{itemize}
    \item We introduce \ours{}, a benchmark of 1,150 expert-curated tasks across eight enterprise domains with outcome-based verification enforcing goal completion, state integrity, policy compliance, and side-effect checks, including 30 infeasible tasks designed to evaluate safe refusal behavior.
    \item We develop a fully interactive, containerized enterprise environment with 164 relational database tables and 512 functional tools, an order of magnitude more complex than prior enterprise benchmarks.
    \item We evaluate 14 frontier models, uncovering and analyzing systematic failure patterns across planning, state management, and policy compliance, and provide actionable insights to build more reliable enterprise agents. We also release \ours{} to the community to advance research in stateful agentic planning and enterprise tool use.
\end{itemize}
\section{Related Works}

\begin{table*}[h]
\centering
\newcommand{\cyes}{{\color{green!60!black}\checkmark}}
\newcommand{\cno}{{\color{red}\xmark}}
\resizebox{\textwidth}{!}{%
\begin{tabular}{lcccccccccc}
\toprule
\textbf{\multirow{2}{*}{Benchmark}}                              & \textbf{\multirow{2}{*}{Focus}} & \textbf{Num.}   & \textbf{Num.}   & \textbf{Num.}   & \textbf{Avg.}   & \textbf{DB}    & \textbf{Avg.} & \textbf{Refusal}  & \textbf{Human Task} & \textbf{Human} \\
                                                                 &                     & \textbf{Domains} & \textbf{Tasks}  & \textbf{Tools}  & \textbf{Steps}  & \textbf{Tables} & \textbf{FK}     & \textbf{Ability?} & \textbf{Curation?}  & \textbf{Plans?} \\
\midrule
\textit{General Tool Use}                                        &                     &                 &                 &                 &                 &                 &                       &                   &                     & \\
API-Bank \citep{li-etal-2023-api}                                & Tool-use             & 8          & 314             & 73              & ~3            & 0                & 0                     & \cno            & \cyes               & \cno \\
ACEBench \citep{chen2025acebenchwinsmatchpoint}                                  & Tool-use             & 8         & 2000            & 4538            & ~2           & 0                & 0                     & \cyes            & \cyes               & \cno \\
$\tau$-bench \citep{yao2024taubenchbenchmarktoolagentuserinteraction}             & User Interaction     & 2         & 165             & 28               & \textemdash             & 3                & 0.7                   & \cyes        & \cyes               & \cno \\
$\tau^2$-bench \citep{barres2025tau2}                                        & User Interaction       & 3                & 279             & 56               & \textemdash             & 9                & \textemdash                   & \cyes        & \cno               & \cno \\
\midrule
\textit{Enterprise Specific}                                     &                     &                 &                 &                 &                 &                 &                       &                   &                     & \\
WorkArena \citep{workarena2024}                                  & ServiceNow          & 7                & 33              & 30             & ~10             & 7                & 0.9                   & \cno            & \cno               & \cno \\
WorkArena++ \citep{boisvert2024workarenacompositionalplanningreasoningbased}     & ServiceNow          & 7                & 682 ($^{\dagger}$341)            & 30             & 30-50           & 7                & \textemdash                   & \cyes        & \cno               & \cno \\
ITBench \citep{jha2025itbenchevaluatingaiagents}                                 & IT                  & 3                & 94              & 10              & \textemdash             & \textemdash              & \textemdash                   & \cno            & \cyes          & \cno \\
WorkBench \citep{styles2024workbenchbenchmarkdatasetagents}                      & Workplace           & 5                & 690 ($^{\dagger}$69)             & 26              & 2             & 5                & 0                     & \cno             & \cyes          & \cno \\
TheAgentCompany \citep{xu2024theagentcompanybenchmarkingllmagents}               & Startup             & 7                & 175             & \textemdash             & \textemdash             & 0              & 0                     & \cno            & \cyes          & \cno \\
CRMArena \citep{huang-etal-2025-crmarena}                                       & Salesforce          & 1                & 1170 ($^{\dagger}$9)             & 27              & \textemdash             & 16               & 1.3                   & \cno            & \cyes          & \cno \\
CRMArena-Pro \citep{huang-etal-2025-crmarena-pro}                                & Salesforce          & 3                & 8560 ($^{\dagger}$19)            & \textemdash             & \textemdash             & 25               & \textemdash                   & \cyes        & \cyes          & \cno \\
EnterpriseBench \citep{vishwakarma2025llmshelpworksandbox}                       & Enterprise          & 5                & 500             & 46              & 3               & 17               & 1.2                   & \cyes        & \cno               & \cno \\ 
\midrule
\rowcolor{gray!15}
\textbf{\ours{} (Ours)}                                          & \textbf{Enterprise} & \textbf{8}       & \textbf{1150}   & \textbf{512}    & \textbf{9$^*$}     & \textbf{164}     & \textbf{1.7}          & \textbf{\cyes}   & \textbf{\cyes}     & \textbf{\cyes} \\
\bottomrule
\end{tabular}%
}
\caption{\textbf{Comparison with existing agentic benchmarks.} \textit{DB Tables} reports the number of unique database tables in the environment; \textit{Avg. FK} measures average foreign keys per table, indicating relational density and the complexity of inter-table dependencies agents must navigate. \textemdash{} denotes values not reported in the original work. $^*$Avg. Steps reflects ideal human-authored execution trajectories; model trajectories may require significantly more steps. Human Plans refer to step-by-step natural language plans written by experts to complete the task. $^{\dagger}$Parenthetical values indicate the number of unique task templates.}
\label{tab:related_works_comparison}
\end{table*}

LLM agent benchmarks broadly fall into three thematic groups: general-purpose tool-use and API evaluation, enterprise platform simulation, and agentic planning and computer-use. \Cref{tab:related_works_comparison} summarizes how \dataset{} compares across key dimensions.

\paragraph{API and Tool-Use Benchmarks}
Early benchmarking efforts focused on testing agents' ability to call APIs and chain tools in general, open-domain settings. ToolLLM \citep{qin2024toolllm} establishes large-scale evaluation across over 16,000 real-world web APIs, while API-Bank \citep{li-etal-2023-api} provides a smaller, runnable system for assessing planning and retrieval across 73 tools. ACEBench \citep{chen2025acebenchwinsmatchpoint} scales this further with 4,538 tools and dynamic multi-turn evaluation, and $\tau$-bench \citep{yao2024taubenchbenchmarktoolagentuserinteraction} and $\tau^2$-bench \citep{barres2025tau2} introduce user simulation and refusal robustness as evaluation axes. These benchmarks make important contributions to measuring tool-calling accuracy but remain anchored to general web-oriented or open-domain APIs. They do not model the multi-system, policy-constrained, stateful tool ecosystems that characterize enterprise environments, which is the setting \ours{} specifically targets.

\paragraph{Enterprise Benchmarks}
A second class of benchmarks targets enterprise platforms directly. Single-platform environments restrict scope to one vendor: WorkArena \citep{workarena2024} and WorkArena++ \citep{boisvert2024workarenacompositionalplanningreasoningbased} evaluate compositional task completion within ServiceNow, CRMArena \citep{huang-etal-2025-crmarena} and CRMArena-Pro \citep{huang-etal-2025-crmarena-pro} focus on Salesforce specific workflows, and ITBench \citep{jha2025itbenchevaluatingaiagents} targets IT incident resolution. Multi-domain efforts broaden the scope but with different emphases: TheAgentCompany \citep{xu2024theagentcompanybenchmarkingllmagents} simulates a startup software company requiring agents to interact via web browser, bash terminal, and code execution, a paradigm fundamentally different from the structured tool-calling API setting of \ours{}; WorkBench \citep{styles2024workbenchbenchmarkdatasetagents} covers office productivity tools; and EnterpriseBench \citep{vishwakarma2025llmshelpworksandbox} evaluates function-calling on sandboxed enterprise data. Across these benchmarks, relational database complexity is consistently limited to fewer than 25 tables, and tasks are either confined to a single vendor ecosystem or lack expert-authored grounding or explicit evaluation of refusal behavior. \ours{} addresses these limitations by spanning eight enterprise domains across both operational systems (CSM, ITSM, HR) and collaboration services (Email, Calendar, Teams, Drive), with 164 database tables with high connectivity, 512 tools, 1,150 expert-curated tasks including policy-constrained infeasible scenarios.

\paragraph{Agentic Planning and Computer-Use}
Planning is a core capability for autonomous agents, and several benchmarks have studied it across diverse general-purpose settings. UserBench \citep{qian2025userbenchinteractivegymenvironment} evaluates agents on multi-turn travel planning tasks where simulated users express preferences incrementally and implicitly, requiring proactive intent elicitation. Gaia2 \citep{froger2025arescalingagentenvironments} evaluates agents in an asynchronous simulated mobile environment, testing search, execution, temporal reasoning, adaptability, ambiguity handling, and multi-agent collaboration. VitaBench \citep{he2025vitabenchbenchmarkingllmagents} evaluates agents on complex multi-turn tasks drawn from real-world services like food delivery, in-store dining, and online travel using a simulated user with dynamic preferences. A parallel line of work focuses on computer-use agents: OSWorld \citep{xie2024osworld}, WindowsAgentArena \citep{bonatti2024arena}, WebArena \citep{zhou2024webarenarealisticwebenvironment} and UI-Vision \citep{nayak2025uivisiondesktopcentricguibenchmark} benchmark agents on desktop and web GUIs requiring complex sequential decision-making. While these settings demand sophisticated planning, they operate in everyday scenarios or computing environments rather than the policy-governed, multi-system contexts of enterprise environments. \ours{} shares the emphasis on extended planning horizons but is uniquely positioned at the intersection of planning depth and enterprise domain fidelity.

\section{\ours{} \raisebox{-.6ex}{\includegraphics[width=0.6cm]{figures/logos/csmgym.png}}\,%
}

In this section, we describe the design and construction of \ours{}, covering domain selection, sandbox environment, task formulation in \Cref{subsec:construction}, and dataset statistics in \Cref{subsec:statistics}.

\begin{figure}[t]
 \centering
 \includegraphics[width=0.7\linewidth]{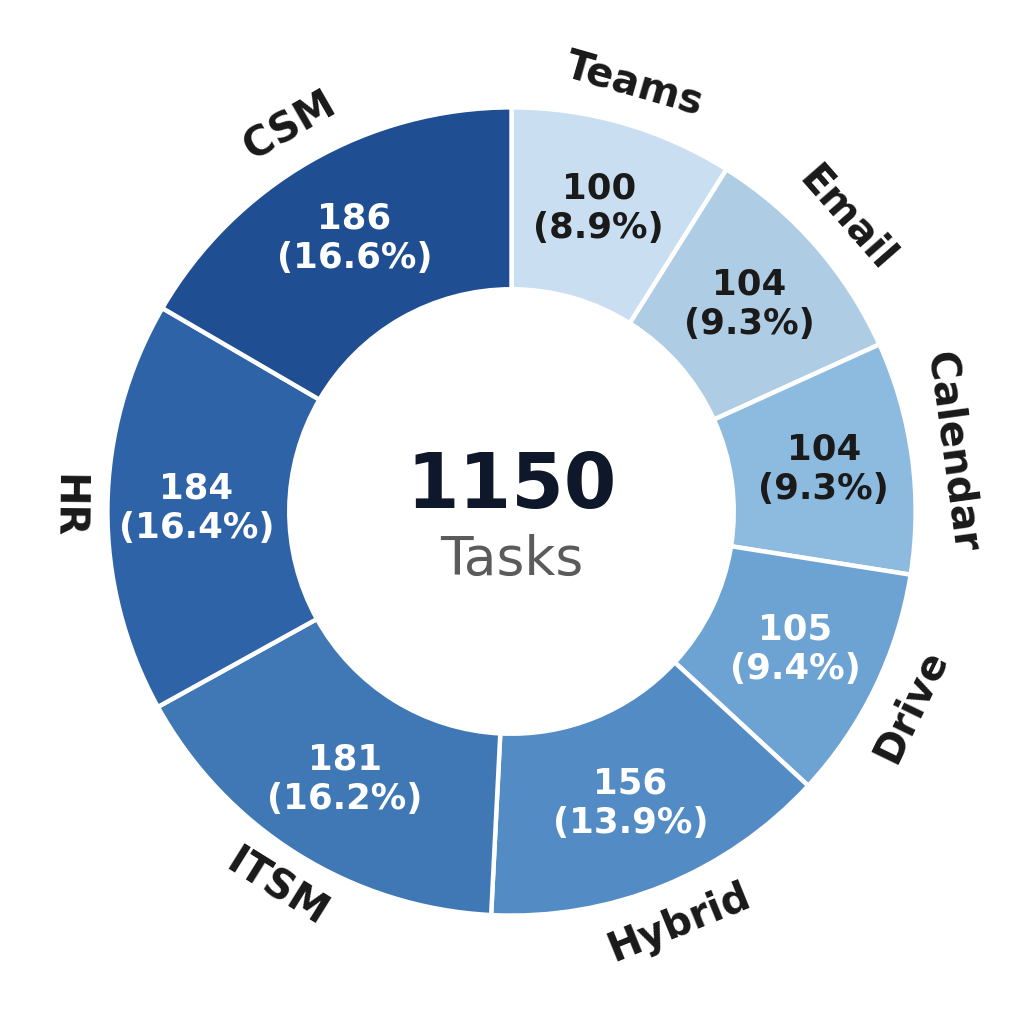}
 \caption{Task distribution across eight \ours{} domains.}
 \label{fig:data_distribution}
\end{figure}

\subsection{\dataset Construction}
\label{subsec:construction}

\paragraph{Selecting domains.}
We selected domains based on three principles, in consultation with SMEs who have hands-on experience in enterprise software: (i) relevance to real-world industry verticals, (ii) diversity in policy complexity and data sensitivity, and (iii) availability of domain experts to author and validate authentic tasks. This led us to two complementary groups of domains.

The first group---\textit{Customer Service Management (CSM)}, \textit{Human Resources (HR)}, and \textit{Information Technology Service Management (ITSM)}---represents the operational backbone of enterprise organizations. These domains are present in virtually every industry vertical and are characterized by strict process compliance, access control policies, and high-stakes workflows where errors have tangible consequences. CSM involves managing the full lifecycle of support tickets and service agreements; HR handles sensitive employee data under strict privacy and procedural rules; and ITSM covers backend IT operations including incident management and system configuration. Their policy-heavy nature and realistic complexity make them ideal for stress-testing constraint-aware planning.

The second group---\textit{Email}, \textit{Calendar}, \textit{Teams}, and \textit{Drive}---encompasses the universal collaboration tools used daily across all enterprise organizations. Agents must be proficient in these to function as effective AI workers. While individually simpler, these domains require sophisticated orchestration: \textit{Email} handles complex mailbox workflows, \textit{Calendar} manages time and resource access policies, \textit{Teams} administers collaborative workspaces, and \textit{Drive} governs file system integrity and security.

Finally, \textit{Hybrid} mandates cross-domain orchestration across these fragmented tools, requiring agents to maintain context and data integrity while switching between systems. Together, the eight domains span the full enterprise workflow spectrum, from internal operations to customer-facing processes to collaboration infrastructure, enabling evaluation of both specialized and general-purpose agentic capabilities. Refer to \Cref{app:categories_sec} for more details on each domain.

\paragraph{Sandbox Environment and Data Generation.}
We partnered with a professional data annotation firm (Turing\footnote{https://www.turing.com}) to assemble a team of over 160 contributors, including software engineers for building environments and Subject Matter Experts (SMEs) for technical domains like CSM, ITSM, and HR (refer to \Cref{app:annot_demographics}). To provide a reproducible and realistic evaluation ground, we developed a containerized docker sandbox hosting domain-specific databases, APIs, and the tool execution layer. This environment mirrors enterprise constraints without exposing proprietary infrastructure. For each domain, we created a realistic database schema and populated it with seed data and tools to access and manipulate data. The underlying data is designed with a strong real-world perspective by SMEs. Guided by official product documentation and SME insights, we model realistic table structures, constraints and tools based on industry standard database schemas. Starting from a fixed seed of data and tools, annotators extend the environment with new tables, schemas, and records as each task demands. Refer to \Cref{app:sandbox} for more details on the environment. 

\paragraph{Task Construction Pipeline.}
The task creation process follows a rigorous pipeline designed to ensure high complexity and faithfulness to real-world workflows. In \textbf{Scenario Design}, annotators craft challenging, multi-step scenarios based on specific complexity thresholds across dimensions including tool invocation counts, verification conditions, and state dependencies (action ordering is crucial to success), access constraints (e.g. ``\textit{only team owners can create private channels}") and other policy conflicts (where user request conflicts with system policies). We constrain tasks to have a unique final state, though multiple valid paths may exist to reach it. In addition, we design a subset of 30 infeasible tasks where completion is intentionally impossible due to insufficient tool availability, explicit policy violations, or resource unavailability. For each task, annotators also update the sandbox environment with any additional necessary tables and tools. We follow this with \textbf{Ground Truth Execution and Plan Authoring}, where for each task, annotators provide a detailed step-by-step plan and manually execute it within the sandbox to capture a gold-standard trajectory, documenting each call with parameters, responses and execution rationale. Annotators also author a natural language reasoning plan that explicitly grounds each action in system constraints, user request, and available tools. Plans reference policies from the system prompt, explaining ordering dependencies thus fully grounding the task in the provided context. Refer to \Cref{app:annot_process} for more details.

We enforce \textbf{Outcome-based Verification}, where annotators author executable SQL verification scripts that check the final state of the environment upon task completion. This ensures that we evaluate agents on \emph{outcomes} rather than rigid action sequences, allowing for alternative valid solution paths. The evaluation includes checks for required conditions (\emph{is the goal achieved?}), integrity constraints (\emph{are foreign key constraints respected?}), permission compliance (\emph{did the agent avoid unauthorized actions?}) and other side effects.  Finally, we conduct multiple rounds of \textbf{Quality Assurance}, where reviewer annotators (including authors) assess task feasibility given the initial state and the available tools,  instruction clarity and completeness without external dependencies or domain knowledge, verification script correctness and coverage, as well as the fluency and coherence of the ground truth plan, instructions and verification scripts. Refer to \Cref{app:annot_qa} for more details on verification.

\subsection{Dataset Statistics} \label{subsec:statistics}

\paragraph{Task Statistics.}
\dataset evaluates agents across 1,150 tasks designed to mimic the depth of real-world enterprise operations, including 30 infeasible tasks that test whether agents correctly refuse unsatisfiable requests. The action space is extensive and diverse, comprising 512 unique tools across domains, with domain-specific toolsets ranging from 37 (Calendar) to 93 (ITSM). Expert human trajectories average 9.15 steps, with planning horizons varying considerably across domains, ranging from 6.2 steps on average in Email to 12.1 in CSM, and reaching up to 34 steps in HR (see \Cref{fig:tasks_len_distribution}). Beyond length, our tasks are dense with constraints: on average, a task mandates satisfying 5.3 distinct verification conditions, with the most intricate scenarios requiring the resolution of 44 conditions.

\paragraph{Environment Statistics.}
Our sandbox environment models a highly interconnected data ecosystem comprising 164 unique database tables across the eight domains. On average, each task interacts with a sub-graph of 24.9 tables, reaching up to 73 in Hybrid scenarios. This means agents must reason over a large, partially-observable data graph to execute each task correctly. Each task operates over an average of 3,443 database rows, scaling to over 10,000 in data-heavy domains like CSM. To quantify relational complexity, we measure the average number of Foreign Keys (FK) per table. We observe a high degree of dependency, with average FKs ranging from 1.1 in Calendar to 2.4 in HR (mean $\approx$ 1.7), exceeding the relational density of prior benchmarks (see \Cref{tab:related_works_comparison}). Higher FK density means agents must resolve more inter-table dependencies when constructing valid tool arguments, making referential integrity a key challenge. We provide more details in \Cref{app:sandbox}.
\section{Experiments}

\begin{table*}[!h]
\centering
\resizebox{\textwidth}{!}{%
\begin{tabular}{lcccccccccc}
\toprule
\textbf{Model} & \textbf{Teams} & \textbf{CSM} & \textbf{Email} & \textbf{ITSM} & \textbf{Calendar} & \textbf{HR} & \textbf{Drive} & \textbf{Hybrid} & \textbf{Infeas.} & \textbf{Avg.}\\
\midrule
\rowcolor{gray!15} \multicolumn{11}{c}{\textit{Closed Source Models}} \\
Claude Opus 4.5 \citep{anthropic2025sonnet45} & 50.0 & 34.2 & {51.9} & 23.8 & \textbf{43.2} & \textbf{32.1} & 49.5 & \textbf{30.7} & 50.0 & \textbf{37.4} \\
Gemini-3-Flash \citep{comanici2025gemini25pushingfrontier} & 47.3 & 35.0 & 44.3 & \textbf{28.5} & 30.5 & 12.6 & 49.7 & 24.2 & 38.5 & 31.9 \\
GPT-5.2 (High) \citep{openai2025gpt5} & 31.0 & 34.8 & 51.0 & 21.7 & 38.5 & 25.0 & 40.0 & 22.2 & 50.0 & 31.8 \\
Claude Sonnet 4.5 \citep{anthropic2025sonnet45} & \textbf{51.0} & 16.7 & 51.3 & 17.6 & 34.6 & 21.6 & \textbf{52.1} & 28.1 & 46.2 & 30.9 \\
GPT-5 \citep{openai2025gpt5} & 26.3 & \textbf{36.4} & 49.0 & 18.9 & 41.3 & 17.9 & 34.0 & 23.5 & 50.5 & 29.8 \\
Gemini-3-Pro \citep{comanici2025gemini25pushingfrontier} & 43.0 & 27.7 & 33.6 & 22.2 & 28.8 & 12.5 & 46.7 & 22.9 & 50.0 & 28.0 \\
GPT-5.2 (Low) \citep{openai2025gpt5} & 25.0 & 21.2 & 43.3 & 6.7 & 28.9 & 13.0 & 26.7 & 20.9 & \textbf{53.9} & 21.9 \\
GPT-5-Mini \citep{openai2025gpt5} & 25.7 & 15.8 & 47.4 & 8.9 & 28.8 & 10.7 & 23.8 & 22.5 & 47.4 & 21.3 \\
Gemini-2.5-Pro \citep{comanici2025gemini25pushingfrontier} & 39.3 & 11.6 & 31.1 & 13.9 & 12.5 & 4.9 & 27.0 & 19.6 & 34.7 & 18.2 \\
\rowcolor{gray!15} \multicolumn{11}{c}{\textit{Open Source Models}} \\
DeepSeek-V3.2 (High)~\citep{deepseekai2024deepseekv3technicalreport} & 37.0 & 14.1 & 47.1 & 16.1 & 21.2 & 16.3 & 35.2 & 22.9 & 53.8 & 24.5 \\
GPT-OSS-120B (High)~\citep{openai2025gptoss120bgptoss20bmodel} & 32.0 & 16.3 & 42.3 & 6.1 & 35.6 & 16.3 & 41.0 & 19.6 & 50.0 & 23.7 \\
DeepSeek-V3.2 (Medium)~\citep{deepseekai2024deepseekv3technicalreport} & 35.7 & 15.4 & 45.8 & 9.6 & 21.5 & 15.0 & 27.6 & 22.9 & 40.0 & 22.3 \\
Kimi-K2-Thinking~\citep{kimiteam2025kimik2openagentic} & 30.0 & 7.1 & 51.0 & 12.2 & 15.4 & 8.2 & 39.6 & 15.7 & 30.5 & 19.5 \\
Qwen3-30B (Think)~\citep{qwen3} & 22.0 & 5.4 & {51.9} & 6.7 & 18.3 & 7.6 & 25.7 & 15.7 & 36.8 & 16.8 \\
Qwen3-235B (Inst.)~\citep{qwen3} & 28.0 & 4.7 & 38.1 & 9.3 & 15.7 & 7.8 & 23.8 & 17.7 & 30.5 & 16.1 \\
Qwen3-4B (Think)~\citep{qwen3} & 24.0 & 3.8 & 38.4 & 5.6 & 5.8 & 7.1 & 21.9 & 15.8 & 31.6 & 13.7 \\
\bottomrule
\end{tabular}%
}
\caption{Overall task completion performance on \ours{}. We report the percentage of tasks successfully completed by each model in oracle tool mode, broken down by domain. A task is considered successful only if all outcome verification checks pass.}
\label{tab:main_results}
\end{table*}

\subsection{Baselines}
\label{subsec:baselines}

We evaluate a diverse set of baselines covering closed-source frontier models, open-source reasoning and non reasoning models. All agents are evaluated under a unified interface with identical task instructions, tool definitions, sandbox environments, and evaluation protocols. Unless stated otherwise, agents operate in an \textit{oracle}-tool setting where we assume a perfect retriever supplies the agent with the right set of tools. This focuses the evaluation purely on planning and execution, without the need for explicit tool discovery. Additionally, we conduct ablations by increasing the number of available tools to analyze how tool set size impacts performance. We use a standard ReAct-style reasoning and tool-execution loop which has been shown to be effective in agentic settings \cite{yao2022react}. The closed-source set includes Claude~4.5 \citep{anthropic2025sonnet45, anthropic2025opus45} variants (Opus and Sonnet), GPT \citep{openai2025gpt5} variants (5.2 High, 5.2 Low, 5, and 5-Mini), and Gemini \citep{comanici2025gemini25pushingfrontier} variants (3-Pro, 3-Flash, and 2.5-Pro), while the open-source set includes Kimi-K2-Thinking \citep{kimiteam2025kimik2openagentic}, DeepSeek-V3.2 \citep{deepseekai2024deepseekv3technicalreport}, GPT-OSS-120B (Medium) \citep{openai2025gptoss120bgptoss20bmodel}, and Qwen3 \citep{qwen3} variants (235B Inst., 30B Think, and 4B Think).

\subsection{Evaluation Metrics}
We evaluate models using pass@1 task completion rate, where a model receives a score of 1 only when it successfully completes all task requirements while satisfying all specified constraints. Task completion is verified by executing SQL-based verifiers hand-written by subject matter experts (SMEs) during the benchmark curation process.
We report the average of pass@1 across three runs (to reduce variance) as our primary metric because it captures end-to-end task success. While we also measure verifier-level success rates (see \Cref{tab:main_results_appendix}), which provide fine-grained insight into the average number of successful verification checks, this metric can be misleading: agents may pass verifiers for trivial trajectory segments (e.g., initial setup steps) while failing on core task logic, system compliance requirements, or side-effect checks. Pass@1 therefore provides a more accurate assessment of real-world agent utility.

\subsection{Results}

\paragraph{How do models perform across different domains?}
Overall, Claude Opus~4.5 achieves the best average task completion (37.4\%) and is particularly strong across several workflows, leading on Email (51.9\%), Calendar (43.2\%), HR (32.1\%), and Hybrid (30.7\%). Gemini-3-Flash emerges as the second-best model overall (31.9\%) and tops ITSM (28.5\%), a service and operations management workflow. Claude Sonnet~4.5 (30.9\%) remains strong on collaboration and document-centric workflows, leading on Teams (51.0\%) and Drive (52.1\%). GPT-5 shows more domain-specific peaks, topping CSM (36.4\%). Open-source models still lag behind the closed-source systems overall. The strongest open-source model, DeepSeek-V3.2 (High), reaches a 24.5\% average, narrowly ahead of GPT-OSS-120B (High) at 23.7\%. They particularly struggle on service, policy, and people-facing domains such as CSM, ITSM, and HR. Qwen3-30B (Think) performs surprisingly well for its size, outperforming its larger instruct variant and attaining a highly competitive Email score (51.9\%, tied for best). Finally, ITSM and Hybrid cross-domain workflows are the hardest settings (best: 28.5\% and 30.7\% respectively), highlighting that service operations and cross-domain coordination remain the key bottlenecks for all model families.

\paragraph{How does adding extra tools affect performance?}
We assessed robustness to tool overload by conducting ablations with Claude Sonnet 4.5, chosen for its strong overall performance and cost-effectiveness. We augmented the oracle toolset with extra \textit{distractor} tools (+5, +10, and +15). To make this setting particularly challenging and representative of realistic retrieval errors, we asked Claude to retrieve the distractor tools that appeared most relevant to the task. Surprisingly, performance remained remarkably stable. The average completion rates actually increased slightly by an average of $\sim$1.0\% (+0.07\% for +5, +2.35\% for +10, and +0.64\% for +15 tools). The only other notable variation was an average 4--9\% increase in output tokens. This suggests that the model utilizes the additional token budget to carefully filter and select the appropriate tools. Such robustness likely stems from the extensive tool-use training inherent in these LLMs. Consequently, our findings indicate that the primary bottleneck for these agents is not tool discovery, but rather task planning and adherence to system policies.

\paragraph{Which model offers the best cost--performance tradeoff?}
Which model offers the best cost--performance trade-off depends on whether the priority is absolute quality or quality per dollar. As shown by the Pareto frontier in \Cref{fig:cost_vs_preformance}, Gemini-3-Flash provides the strongest practical trade-off among closed-source models. It achieves 31.9\% performance at 0.03 USD per task, delivering a higher success rate than more expensive models like GPT-5 (29.8\% at 0.16 USD) and Claude Sonnet 4.5 (30.9\% at 0.26 USD) at a fraction of the cost (80--90\% less). Within the open-source ecosystem, DeepSeek-V3.2 (High) emerges as the Pareto-dominant option, achieving 24.5\% performance at just 0.014 USD closely followed by GPT-OSS-120B (High) (23.7\% at 0.015 USD), making these the best open-source value overall. Qwen3-235B (Inst.) remains the cheapest overall option (0.007 USD) but comes with a significant performance floor of 16.1\%. Given that success rates across the board remain below 40\%, these systems are not yet reliable enough for autonomous deployment without human oversight. For the highest absolute reliability, Claude Opus 4.5 remains the premier choice (37.4\%), though it requires a steep premium of 0.36 USD per task.

\paragraph{How well do models refuse infeasible tasks?}
We curate 30 infeasible tasks across the 8 domains to evaluate whether models appropriately abstain from unsatisfiable requests. Each task has an average of 10 verification checks to ensure there are no side effects on the system. Tasks are impossible through three primary mechanisms: insufficient tool availability, explicit policy violations (e.g. scheduling conflicts, data access rules etc) and resource unavailability (e.g. inactive users, system in migration mode etc). Moreover, tasks employ compound constraints averaging 3 to 4 per request, necessitating that models evaluate multiple intersecting conditions to identify task feasibility. As seen in Table~\ref{tab:main_results}, GPT-5.2 (Low) and DeepSeek-V3.2 (High) perform the best (53.9\% and 53.8\% respectively) in abstaining from the task while leaving no side effects on the system, followed closely by GPT-5 (50.5\%) and a cluster of models including Claude Opus 4.5, Gemini-3-Pro, and GPT-OSS-120B (High) at 50.0\%. However, the absolute scores of all the models remain well below safe applicability for production systems.

\begin{figure}[t]
 \centering
 \includegraphics[width=1.0\linewidth]{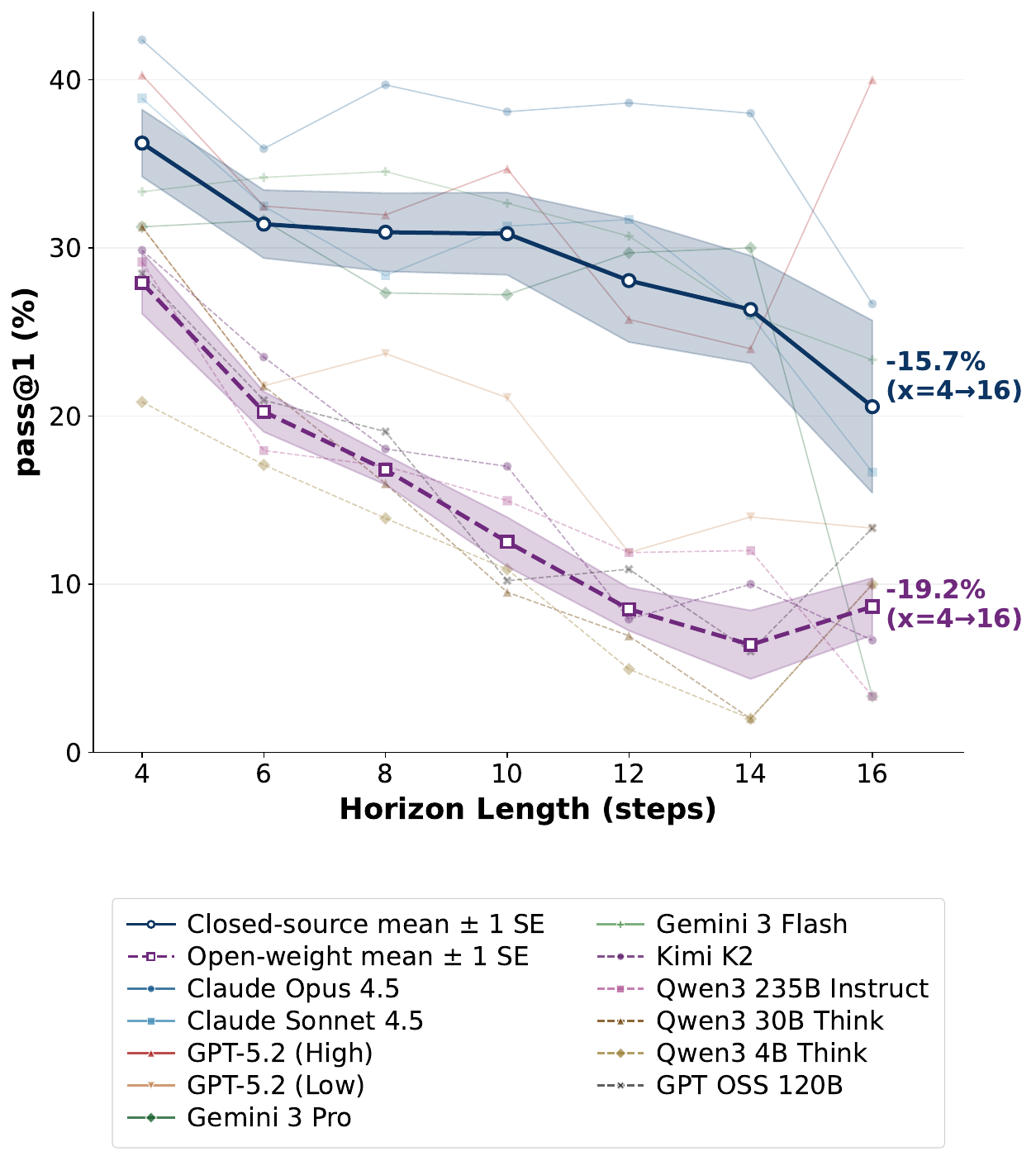}
 \caption{\textbf{Performance degrades consistently with planning horizon.}
Pass@1 accuracy for closed-source (solid) and open-weight (dashed) models
across horizon lengths 4–16. Thick lines show the group mean $\pm$1 SE.
We observe monotonic degradation of performance for both sets, while open model performance falls more sharply with horizon length.}
 \label{fig:performance_by_horizon}
\end{figure}

\begin{figure*}[t]
 \centering
 \includegraphics[width=1.0\linewidth]{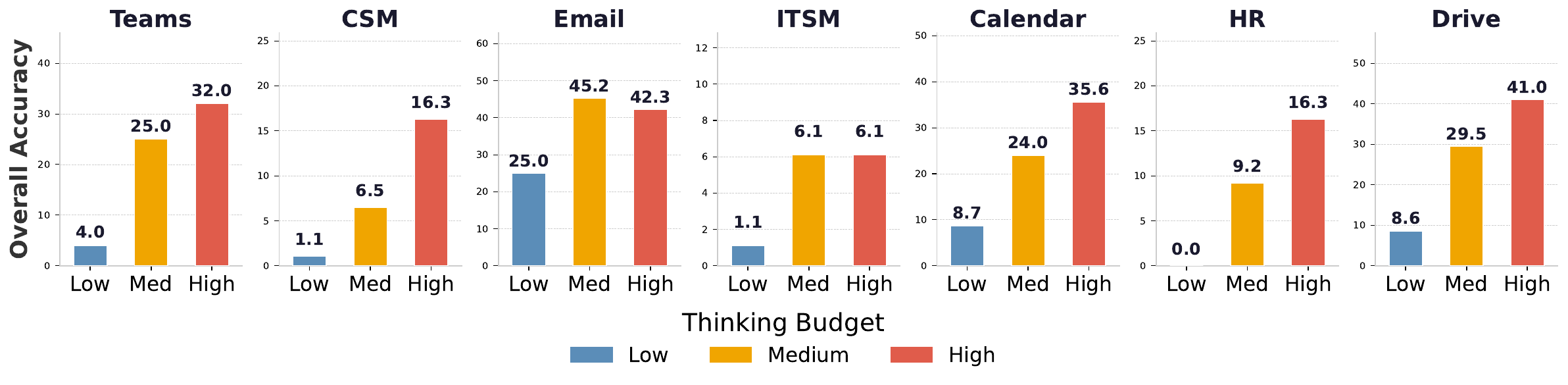}
 \caption{\textbf{Impact of thinking budget on performance}
Histograms show the performance numbers with thinking budget with GPT-OSS-120B model \citep{openai2025gptoss120bgptoss20bmodel} across domains. The results show that the model with \textit{low} thinking budget performing poorly with performance steadily increasing with thinking budget. }
 \label{fig:thinking_budget_ablation}
\end{figure*}

\paragraph{How does performance scale with task horizon?}
To understand how model capabilities degrade with task complexity, we stratify tasks by their expected horizon length, which is proportional to the number of tool execution steps in the human created execution trajectories. As illustrated in \Cref{fig:performance_by_horizon}, performance across all models exhibits a consistent decay as the task horizon increases, reflecting the cumulative difficulty of maintaining reasoning integrity over multi-step sequences. The closed-source group, led by Claude Opus 4.5, demonstrates greater resilience, maintaining a performance lead even as the group mean drops from approximately 35\% at 4 steps to under 20\% by step 16. In contrast, the open-source cohort shows a much steeper decline, with models like Kimi K2 and GPT OSS 120B converging toward a success rate near 10\% at the maximum horizon. This near-universal trend suggests that while current models can navigate short-to-medium sequences, the rapid accumulation of errors in long-horizon tasks remains a critical barrier to autonomous reliability in production environments.

\paragraph{How does thinking budget affect performance?}
We evaluated the impact of test-time compute by varying the thinking budget (\textit{low, medium, high}) for the GPT-OSS-120B model \citep{openai2025gptoss120bgptoss20bmodel}. Increasing the thinking budget yields substantial improvements in task completion across almost all domains (see Figure~\ref{fig:thinking_budget_ablation}). Operating at a \textit{low} thinking budget causes the model to struggle severely, achieving near-zero accuracy on complex service and people-facing domains such as CSM (1.1\%), ITSM (1.1\%), and HR (0.0\%). Scaling to a \textit{high} budget unlocks significant capabilities, driving dramatic absolute gains in Drive ($8.6 \to 41.0$\%), Calendar ($8.7 \to 35.6$\%), and Teams ($4.0 \to 32.0$\%). This strong dependence on test-time compute underscores that \ours{} tasks require complex reasoning and planning to execute. However, we also observe that performance scaling is not universally monotonic; for instance, performance on Email peaks at the \textit{medium} budget (45.2\%) before receding slightly, and ITSM plateaus early ($1.1 \to 6.1\% \to 6.1\%$). This suggests that simply allocating more thinking tokens cannot universally overcome fundamental capability bottlenecks in certain workflows.

\subsection{Further Analysis}

\newcommand{\up}[1]{\textsubscript{\color{green!50!black}$\uparrow$#1\%}}

\begin{table}[ht]
\centering
\small
\begin{tabular}{@{} l l S[table-format=2.2] @{\,} l S[table-format=2.1] @{\,} l S[table-format=2.2] @{\,} l @{}}
\toprule
\textbf{Model} & \textbf{Plan} & \multicolumn{2}{c}{\textbf{CSM}} & \multicolumn{2}{c}{\textbf{ITSM}} & \multicolumn{2}{c}{\textbf{HR}} \\
\midrule
\multirow{2}{*}{Kimi-K2}   & CP & 19.6  & \up{12.5} & 18.1 & \up{5.9}  & 17.2 & \up{9.0} \\
                           & HP & 42.2  & \up{35.1} & 29.1 & \up{16.9} & 34.5 & \up{26.3} \\ 
\cmidrule(lr){1-8}
\multirow{2}{*}{Qwen3-30B} & CP & 15.2  & \up{9.8}  & 11.7 & \up{5.0}  & 17.9 & \up{10.3} \\
                           & HP & 33.9  & \up{28.5} & 20.9 & \up{14.2} & 33.2 & \up{25.6} \\ 
\cmidrule(lr){1-8}
\multirow{2}{*}{Qwen3-4B}  & CP & 16.8  & \up{13.0} & 12.2 & \up{6.6}  & 19.6 & \up{12.5} \\
                           & HP & 37.16 & \up{33.4} & 23.3 & \up{17.7} & 36.4 & \up{29.3} \\
\bottomrule
\end{tabular}
\caption{Plan-Conditioned Execution Baseline. Comparison of performance with Claude Plans (CP) vs. Human Plans (HP). Green values indicate \% improvement over the ReAct baseline from Table~\ref{tab:main_results}.}
\label{tab:plan_conditioned}
\end{table}

The results above show that performance degrades sharply with horizon length and that models struggle most with planning rather than execution. To understand this further, we run a series of controlled ablations that both decouple planning from execution and build increasingly complex multi-agent systems to test whether distributing cognitive load across specialized agents can recover performance that a single monolithic agent cannot achieve.

\textbf{Automated planning improves weaker models consistently.} We introduce a planner-executor baseline in which a dedicated planner agent, instantiated with Claude Sonnet 4.5, one of our best peforming model, generates a high-level plan reasoning over user intent, policy constraints, and potential side effects. A separate executor then carries out tool execution using the same ReAct loop as the single-agent baseline. We evaluate three weaker models (Kimi-K2, Qwen-30B, and Qwen-4B) on the three most challenging domains (CSM, ITSM, and HR). As shown in Table~\ref{tab:plan_conditioned}, performance improves consistently across all models and domains with gains of 6-13\% confirming that planning quality is a meaningful bottleneck even when the executor model is fixed. 

\begin{figure}[h!]
 \centering
 \includegraphics[width=1.0\linewidth]{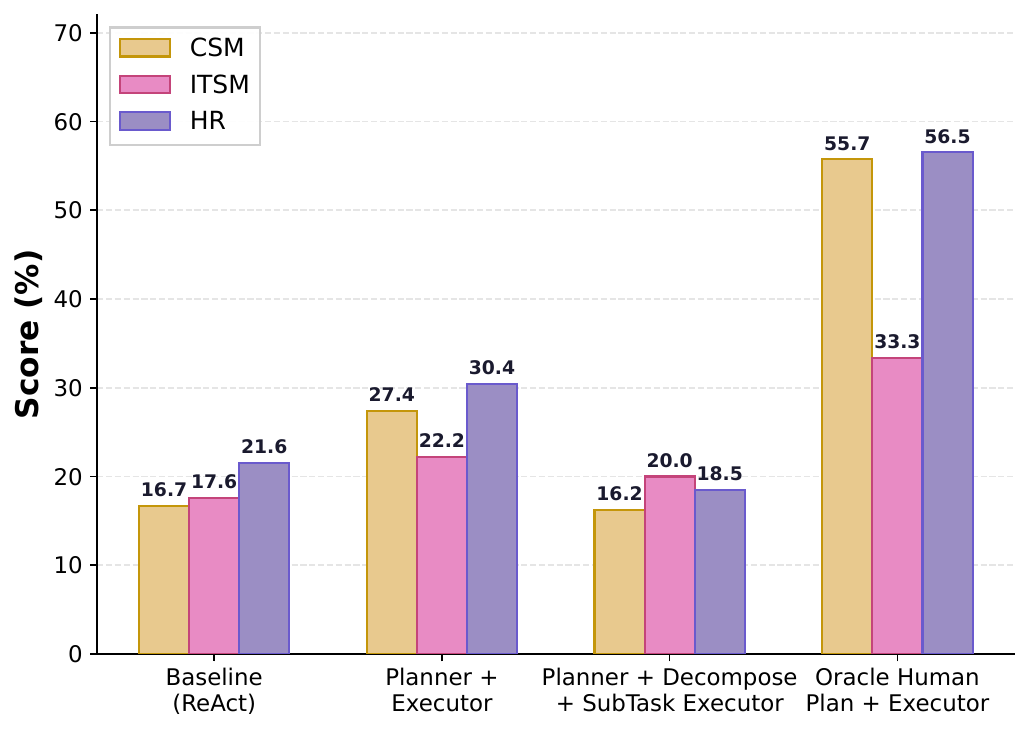}
 \caption{\textbf{ Impact of agentic orchestration on performance}
Histograms show the performance numbers with various multi-agentic architectures using Claude-Sonnet-4.5. The baseline is the simple ReAct loop described in Section~\ref{subsec:baselines}. \textit{Planner+Executor} architecture first prompts the model for a detailed plan, and performs a ReAct loop conditioned on the plan. \textit{Planner+Decompose+Subtask Executor} additionally does a task decomposition and calls subagents in a ReAct loop for each subtask. Finally, \textit{Oracle Human Plan + Executor} performs a ReAct execution loop conditioned on a human written plan.} 
 \label{fig:mas_orchestration_analysis}
\end{figure}

\textbf{Human-authored plans reveal a substantially higher ceiling.} To bound what better planning could achieve, we give the same executor models human-authored reference plans and ask them to carry out the corresponding tool execution, fully decoupling planning from execution. The gains are considerably larger than those from automated planning: 14-35 percentage points across models and domains, roughly doubling the improvements seen from Claude-generated plans. Notably, Qwen3-4B with human-authored plans (and Claude plans) is competitive with or outperforms larger models under the same condition. This suggests that when strategic reasoning is externalized, the primary remaining challenge is faithful instruction-following and precise tool invocation, both capabilities in which modern LLMs exhibit broad competence regardless of scale. This may also indicate that larger models, having stronger internal priors, are more prone to deviating from a provided plan, while smaller models tend to follow step-by-step instructions more literally---an advantage when those instructions are optimal. However, gap between the human plans and LLM plans indicates that current LLMs fall well short of human-level strategic reasoning on these tasks.

\textbf{More complex orchestration does not close this gap.} We evaluate two MAS configurations for Claude Sonnet 4.5: a \textit{Planner+Executor} system that conditions ReAct on an auto-generated plan, and a \textit{Planner+Decompose+Subtask Executor} system that additionally decomposes tasks and invokes a separate subagent per subtask. We evaluate these systems against the ReAct executor baseline with and without human authored plans. As shown in Figure~\ref{fig:mas_orchestration_analysis}, the \textit{Planner+Executor} setup consistently outperforms the ReAct baseline, yielding absolute gains of 10.7\% in CSM and 8.8\% in HR. However, the decomposition architecture is less robust. While it provides a minor lift in ITSM, it regresses in both CSM and HR, even falling below the base ReAct performance in CSM (16.2\% vs. 16.7\%). This is consistent with \ours{} tasks having strong sequential state dependencies that decomposition disrupts. Ultimately, the substantial remaining gap between automated systems and ReAct with human plans suggests that progress requires advances in constraint-aware plan generation rather than architectural complexity alone.

\paragraph{Failure Modes}
We performed a manual qualitative analysis of samples where models made partial progress but ultimately failed to complete the task. We observe several recurring failure patterns. Models frequently invoke tools that create database objects without first querying the necessary prerequisites, producing dangling records with broken foreign-key links \textbf{(Missing Prerequisite Lookup)}. For example, in a task requiring the creation of an HR topic under a specific category, the model skips retrieving available categories and inserts an orphaned record. Models also fail to trigger the follow-up actions mandated by system policies when certain state transitions occur \textbf{(Cascading State Propagation)}. Further failure modes include passing unverified identifiers to tool calls instead of resolving the correct IDs through prior tool interactions \textbf{(Incorrect ID Resolution)}, and prematurely declaring task completion before all required steps have been executed \textbf{(Premature Completion Hallucination)}. To futher systematically categorize these errors, we use an LLM (Claude Sonnet 4.5~\cite{claudeopus}) to tag the final-state SQL verifiers into three types based on their expert-written descriptions: i)~\textit{Task Completion} verifiers check whether the primary user objective was achieved; ii)~\textit{Integrity Constraints} verifiers check that the system remains in a consistent state with valid foreign-key relationships; and iii)~\textit{Permission and Process Compliance} verifiers check adherence to system policies governing permissions and procedural rules. Verifier pass rates for these categories are reported in \Cref{tab:verifier_analysis}. Models struggle most with \textit{Permission and Process Compliance}---a particularly critical gap for real-world deployment, where policy violations can cause cascading system failures and introduce serious security vulnerabilities. Refer to \Cref{app:errors} for examples of model failures.

\section{Discussion and Conclusion}

We introduced \ours{}, a benchmark and sandboxed evaluation platform spanning 1,150 expert-curated tasks across eight enterprise productivity domains, with 512 tools and SQL-based verifiers authored by subject matter experts. Our experiments surface several findings that we believe have broad implications for the development of enterprise-grade LLM agents.

\textbf{Current agents are far from enterprise-ready.}
Even under oracle tool access---the most favorable possible retrieval setting---the best model achieves only 37.4\% task success, and performance degrades monotonically with horizon length and cross-domain coupling. Critically, failure modes are not random: our qualitative and quantitative analysis shows that agents struggle most with \textit{Permission and Process Compliance} (e.g., policy adherence, cascading state transitions) rather than with basic task completion. Furthermore, even frontier models refuse infeasible tasks reliably only about half the time (best: 53.9\%), falling well short of the robustness required for unsupervised deployment. These results indicate that the gap to enterprise reliability will not be closed by scaling model capacity alone.

\textbf{Planning is the dominant bottleneck, not tool execution.}
Our plan-conditioned ablations demonstrate that human-authored plans yield 14--35 percentage point gains across models and domains which is far larger than gains from automated planning or more complex multi-agent orchestration. Strikingly, Qwen3-4B conditioned on human plans is competitive with much larger models under the same condition, suggesting that once strategic reasoning is externalized, even small models can execute faithfully. This dissociation between planning and execution ability implies that the core challenge is constraint-aware plan generation, not tool invocation proficiency. Consistent with this, distractor tools do not meaningfully hurt performance further confirming that tool retrieval is not the binding constraint. Advances in long-horizon, policy-aware planning are therefore the highest-leverage direction for improving agent performance on \ours{}.

\textbf{Thinking budget matters, but has domain-specific ceilings.}
Increasing test-time compute yields substantial gains in most domains, but scaling is not universally monotonic: some domains plateau early, suggesting that additional reasoning tokens cannot compensate for fundamental gaps in domain knowledge or policy understanding. Future work should investigate how to allocate test-time compute more adaptively, and whether targeted training on constraint-heavy domains can raise these ceilings.

\textbf{Future directions.}
Our results motivate three concrete research priorities. First, \textit{constraint-aware plan generation}: methods that explicitly reason over policy constraints, side-effect dependencies, and prerequisite structures before committing to action sequences. Second, \textit{long-horizon state management}: mechanisms for maintaining coherent world state over many tool calls, such as episodic memory or structured state representations, to prevent the error accumulation we observe with increasing horizon length. Third, \textit{safe refusal and escalation}: agents must reliably detect infeasible or policy-violating requests and abstain cleanly, a capability that remains weak across all evaluated systems today.

We will release \ours{}, its sandbox environment, and evaluation tooling to support open, community-driven research. The sandbox is modular and extensible, allowing new domains, tools, and workflows to be added as enterprise practices evolve. By grounding agent evaluation in realistic, constraint-rich enterprise workflows, \ours{} aims to shift the field's focus toward the planning, safety, and policy-compliance capabilities that truly determine whether an LLM agent is deployable as a reliable \textit{AI worker}.

\section{Acknowledgments}
We gratefully acknowledge Turing as our data curation partner for this project. We extend special thanks to Ankit Jasuja, Aakash Chavan, Harshil Parekh, Anuj Jain, Igor Vidal, Rahul Bora, and Sudarshan Sivaraman from Turing, whose instrumental contributions to sandbox engineering and sample generation upheld the highest standards of quality throughout. We also thank the more than 160 data contributors and engineers who painstakingly authored benchmark samples under specific guidance and iterative feedback from the authors — their dedication was indispensable to the scale and fidelity of this benchmark.
This work further benefited from the meticulous quality assurance oversight provided by the linguistics team at ServiceNow, particularly Racheal Hansen, Tiffany Do, and Nidhi Kumari. We also thank Rabiul Awal for his helpful feedback on an early draft of this paper. Finally, we gratefully acknowledge Patrice Bechard and Vikas Yadav at ServiceNow for their valuable feedback throughout the development of this work.


\bibliography{icml2026}

@misc{barres2025tau2,
      title={$\tau^2$-Bench: Evaluating Conversational Agents in a Dual-Control Environment}, 
      author={Victor Barres and Honghua Dong and Soham Ray and Xujie Si and Karthik Narasimhan},
      year={2025},
      eprint={2506.07982},
      archivePrefix={arXiv},
      primaryClass={cs.AI},
      url={https://arxiv.org/abs/2506.07982}, 
}

@inproceedings{li-etal-2023-api,
    title = "{API}-Bank: A Comprehensive Benchmark for Tool-Augmented {LLM}s",
    author = "Li, Minghao  and
      Zhao, Yingxiu  and
      Yu, Bowen  and
      Song, Feifan  and
      Li, Hangyu  and
      Yu, Haiyang  and
      Li, Zhoujun  and
      Huang, Fei  and
      Li, Yongbin",
    editor = "Bouamor, Houda  and
      Pino, Juan  and
      Bali, Kalika",
    booktitle = "Proceedings of the 2023 Conference on Empirical Methods in Natural Language Processing",
    month = dec,
    year = "2023",
    address = "Singapore",
    publisher = "Association for Computational Linguistics",
    url = "https://aclanthology.org/2023.emnlp-main.187/",
    doi = "10.18653/v1/2023.emnlp-main.187",
    pages = "3102--3116"
}

@inproceedings{qin2024toolllm,
    title={Tool{LLM}: Facilitating Large Language Models to Master 16000+ Real-world {API}s},
    author={Yujia Qin and Shihao Liang and Yining Ye and Kunlun Zhu and Lan Yan and Yaxi Lu and Yankai Lin and Xin Cong and Xiangru Tang and Bill Qian and Sihan Zhao and Lauren Hong and Runchu Tian and Ruobing Xie and Jie Zhou and Mark Gerstein and dahai li and Zhiyuan Liu and Maosong Sun},
    booktitle={The Twelfth International Conference on Learning Representations},
    year={2024},
    url={https://openreview.net/forum?id=dHng2O0Jjr}
}

@misc{chen2025acebenchwinsmatchpoint,
      title={ACEBench: Who Wins the Match Point in Tool Usage?}, 
      author={Chen Chen and Xinlong Hao and Weiwen Liu and Xu Huang and Xingshan Zeng and Shuai Yu and Dexun Li and Shuai Wang and Weinan Gan and Yuefeng Huang and Wulong Liu and Xinzhi Wang and Defu Lian and Baoqun Yin and Yasheng Wang and Wu Liu},
      year={2025},
      eprint={2501.12851},
      archivePrefix={arXiv},
      primaryClass={cs.CL},
      url={https://arxiv.org/abs/2501.12851}, 
}

@misc{workarena2024,
      title={WorkArena: How Capable Are Web Agents at Solving Common Knowledge Work Tasks?}, 
      author={Alexandre Drouin and Maxime Gasse and Massimo Caccia and Issam H. Laradji and Manuel Del Verme and Tom Marty and Léo Boisvert and Megh Thakkar and Quentin Cappart and David Vazquez and Nicolas Chapados and Alexandre Lacoste},
      year={2024},
      eprint={2403.07718},
      archivePrefix={arXiv},
      primaryClass={cs.LG}
}

@misc{boisvert2024workarenacompositionalplanningreasoningbased,
      title={WorkArena++: Towards Compositional Planning and Reasoning-based Common Knowledge Work Tasks}, 
      author={Léo Boisvert and Megh Thakkar and Maxime Gasse and Massimo Caccia and Thibault Le Sellier De Chezelles and Quentin Cappart and Nicolas Chapados and Alexandre Lacoste and Alexandre Drouin},
      year={2024},
      eprint={2407.05291},
      archivePrefix={arXiv},
      primaryClass={cs.AI},
      url={https://arxiv.org/abs/2407.05291}, 
}

@inproceedings{huang-etal-2025-crmarena,
    title = "CRMArena: Understanding the Capacity of LLM Agents to Perform Professional CRM Tasks in Realistic Environments",
    author = "Huang, Kung-Hsiang  and
      Prabhakar, Akshara  and
      Dhawan, Sidharth  and
      Mao, Yixin  and
      Wang, Huan  and
      Savarese, Silvio  and
      Xiong, Caiming  and
      Laban, Philippe  and
      Wu, Chien-Sheng",
    booktitle = "Proceedings of the 2025 Conference of the Nations of the Americas Chapter of the Association for Computational Linguistics: Human Language Technologies (Volume 1: Long Papers)",
    year = "2025",
}

@article{huang-etal-2025-crmarena-pro,
    title = "CRMArena-Pro: Holistic Assessment of LLM Agents Across Diverse Business Scenarios and Interactions",
    author = "Huang, Kung-Hsiang  and
      Prabhakar, Akshara  and
      Thorat, Onkar  and
      Agarwal, Divyansh  and
      Choubey, Prafulla Kumar  and
      Mao, Yixin  and
      Savarese, Silvio  and
      Xiong, Caiming  and
      Wu, Chien-Sheng",
    journal = "arXiv preprint arXiv:2505.18878",
    year = "2025",
}

@misc{jha2025itbenchevaluatingaiagents,
      title={ITBench: Evaluating AI Agents across Diverse Real-World IT Automation Tasks}, 
      author={Saurabh Jha and Rohan Arora and Yuji Watanabe and Takumi Yanagawa and Yinfang Chen and Jackson Clark and Bhavya Bhavya and Mudit Verma and Harshit Kumar and Hirokuni Kitahara and Noah Zheutlin and Saki Takano and Divya Pathak and Felix George and Xinbo Wu and Bekir O. Turkkan and Gerard Vanloo and Michael Nidd and Ting Dai and Oishik Chatterjee and Pranjal Gupta and Suranjana Samanta and Pooja Aggarwal and Rong Lee and Pavankumar Murali and Jae-wook Ahn and Debanjana Kar and Ameet Rahane and Carlos Fonseca and Amit Paradkar and Yu Deng and Pratibha Moogi and Prateeti Mohapatra and Naoki Abe and Chandrasekhar Narayanaswami and Tianyin Xu and Lav R. Varshney and Ruchi Mahindru and Anca Sailer and Laura Shwartz and Daby Sow and Nicholas C. M. Fuller and Ruchir Puri},
      year={2025},
      eprint={2502.05352},
      archivePrefix={arXiv},
      primaryClass={cs.AI},
      url={https://arxiv.org/abs/2502.05352}, 
}

@misc{styles2024workbenchbenchmarkdatasetagents,
      title={WorkBench: a Benchmark Dataset for Agents in a Realistic Workplace Setting}, 
      author={Olly Styles and Sam Miller and Patricio Cerda-Mardini and Tanaya Guha and Victor Sanchez and Bertie Vidgen},
      year={2024},
      eprint={2405.00823},
      archivePrefix={arXiv},
      primaryClass={cs.CL},
      url={https://arxiv.org/abs/2405.00823}, 
}

@misc{vishwakarma2025llmshelpworksandbox,
      title={Can LLMs Help You at Work? A Sandbox for Evaluating LLM Agents in Enterprise Environments}, 
      author={Harsh Vishwakarma and Ankush Agarwal and Ojas Patil and Chaitanya Devaguptapu and Mahesh Chandran},
      year={2025},
      eprint={2510.27287},
      archivePrefix={arXiv},
      primaryClass={cs.LG},
      url={https://arxiv.org/abs/2510.27287}, 
}

@misc{xu2024theagentcompanybenchmarkingllmagents,
      title={TheAgentCompany: Benchmarking LLM Agents on Consequential Real World Tasks}, 
      author={Frank F. Xu and Yufan Song and Boxuan Li and Yuxuan Tang and Kritanjali Jain and Mengxue Bao and Zora Z. Wang and Xuhui Zhou and Zhitong Guo and Murong Cao and Mingyang Yang and Hao Yang Lu and Amaad Martin and Zhe Su and Leander Maben and Raj Mehta and Wayne Chi and Lawrence Jang and Yiqing Xie and Shuyan Zhou and Graham Neubig},
      year={2024},
      eprint={2412.14161},
      archivePrefix={arXiv},
      primaryClass={cs.CL},
      url={https://arxiv.org/abs/2412.14161}, 
}

@article{yao2022react,
  title     = {ReAct: Synergizing Reasoning and Acting in Language Models},
  author    = {Yao, Shunyu and Zhao, Jeffrey and Yu, Dian and Du, Nan and Shafran, Izhak and Narasimhan, Karthik and Cao, Yuan},
  journal   = {arXiv preprint arXiv:2210.03629},
  year      = {2022}
}

@misc{claudeopus,
  title     = {Introducing Claude Opus 4.5},
  author    = {Anthropic},
  year      = {2025}
}

@article{xie2024osworld,
  title        = {OSWorld: Benchmarking Multimodal Agents for Open-Ended Tasks in Real Computer Environments},
  author       = {Xie, Tianbao and others},
  journal      = {arXiv preprint arXiv:2404.07972},
  year         = {2024}
}

@misc{kimiteam2025kimik2openagentic,
      title={Kimi K2: Open Agentic Intelligence}, 
      author={K-Team},
      year={2025},
      eprint={2507.20534},
      archivePrefix={arXiv},
      primaryClass={cs.LG},
      url={https://arxiv.org/abs/2507.20534}, 
}

@misc{deepseekai2024deepseekv3technicalreport,
      title={DeepSeek-V3 Technical Report}, 
      author={DeepSeek-AI},
      year={2024},
      eprint={2412.19437},
      archivePrefix={arXiv},
      primaryClass={cs.CL},
      url={https://arxiv.org/abs/2412.19437}, 
}

@article{qwen3,
    title={Qwen3 Technical Report}, 
    author={An Yang and Anfeng Li and Baosong Yang and Beichen Zhang and Binyuan Hui and Bo Zheng and Bowen Yu and Chang Gao and Chengen Huang and Chenxu Lv and Chujie Zheng and Dayiheng Liu and Fan Zhou and Fei Huang and Feng Hu and Hao Ge and Haoran Wei and Huan Lin and Jialong Tang and Jian Yang and Jianhong Tu and Jianwei Zhang and Jianxin Yang and Jiaxi Yang and Jing Zhou and Jingren Zhou and Junyang Lin and Kai Dang and Keqin Bao and Kexin Yang and Le Yu and Lianghao Deng and Mei Li and Mingfeng Xue and Mingze Li and Pei Zhang and Peng Wang and Qin Zhu and Rui Men and Ruize Gao and Shixuan Liu and Shuang Luo and Tianhao Li and Tianyi Tang and Wenbiao Yin and Xingzhang Ren and Xinyu Wang and Xinyu Zhang and Xuancheng Ren and Yang Fan and Yang Su and Yichang Zhang and Yinger Zhang and Yu Wan and Yuqiong Liu and Zekun Wang and Zeyu Cui and Zhenru Zhang and Zhipeng Zhou and Zihan Qiu},
    journal = {arXiv preprint arXiv:2505.09388},
    year={2025}
}

@misc{nayak2025uivisiondesktopcentricguibenchmark,
  title={UI-Vision: A Desktop-centric GUI Benchmark for Visual Perception and Interaction}, 
  author={Shravan Nayak and Xiangru Jian and Kevin Qinghong Lin and Juan A. Rodriguez and 
          Montek Kalsi and Rabiul Awal and Nicolas Chapados and M. Tamer Özsu and 
          Aishwarya Agrawal and David Vazquez and Christopher Pal and Perouz Taslakian and 
          Spandana Gella and Sai Rajeswar},
  year={2025},
  eprint={2503.15661},
  archivePrefix={arXiv},
  primaryClass={cs.CV},
  url={https://arxiv.org/abs/2503.15661}, 
}

@inproceedings{
    jimenez2024swebench,
    title={{SWE}-bench: Can Language Models Resolve Real-world Github Issues?},
    author={Carlos E Jimenez and John Yang and Alexander Wettig and Shunyu Yao and Kexin Pei and Ofir Press and Karthik R Narasimhan},
    booktitle={The Twelfth International Conference on Learning Representations},
    year={2024},
    url={https://openreview.net/forum?id=VTF8yNQM66}
}

@misc{yao2024taubenchbenchmarktoolagentuserinteraction,
      title={$\tau$-bench: A Benchmark for Tool-Agent-User Interaction in Real-World Domains}, 
      author={Shunyu Yao and Noah Shinn and Pedram Razavi and Karthik Narasimhan},
      year={2024},
      eprint={2406.12045},
      archivePrefix={arXiv},
      primaryClass={cs.AI},
      url={https://arxiv.org/abs/2406.12045}, 
}

@misc{qian2025userbenchinteractivegymenvironment,
      title={UserBench: An Interactive Gym Environment for User-Centric Agents}, 
      author={Cheng Qian and Zuxin Liu and Akshara Prabhakar and Zhiwei Liu and Jianguo Zhang and Haolin Chen and Heng Ji and Weiran Yao and Shelby Heinecke and Silvio Savarese and Caiming Xiong and Huan Wang},
      year={2025},
      eprint={2507.22034},
      archivePrefix={arXiv},
      primaryClass={cs.AI},
      url={https://arxiv.org/abs/2507.22034}, 
}

@misc{froger2025arescalingagentenvironments,
      title={ARE: Scaling Up Agent Environments and Evaluations}, 
      author={Romain Froger and Pierre Andrews and Matteo Bettini and Amar Budhiraja and Ricardo Silveira Cabral and Virginie Do and Emilien Garreau and Jean-Baptiste Gaya and Hugo Laurençon and Maxime Lecanu and Kunal Malkan and Dheeraj Mekala and Pierre Ménard and Gerard Moreno-Torres Bertran and Ulyana Piterbarg and Mikhail Plekhanov and Mathieu Rita and Andrey Rusakov and Vladislav Vorotilov and Mengjue Wang and Ian Yu and Amine Benhalloum and Grégoire Mialon and Thomas Scialom},
      year={2025},
      eprint={2509.17158},
      archivePrefix={arXiv},
      primaryClass={cs.AI},
      url={https://arxiv.org/abs/2509.17158}, 
}

@inproceedings{
bonatti2024arena,
title={Windows Agent Arena: Evaluating Multi-Modal {OS} Agents at Scale},
author={Rogerio Bonatti and Dan Zhao and Francesco Bonacci and Dillon Dupont and Sara Abdali and Yinheng Li and Yadong Lu and Justin Wagle and Kazuhito Koishida and Arthur Bucker and Lawrence Keunho Jang and Zheng Hui},
booktitle={Forty-second International Conference on Machine Learning},
year={2025},
url={https://openreview.net/forum?id=W9s817KqYf}
}

@misc{he2025vitabenchbenchmarkingllmagents,
      title={VitaBench: Benchmarking LLM Agents with Versatile Interactive Tasks in Real-world Applications}, 
      author={Wei He and Yueqing Sun and Hongyan Hao and Xueyuan Hao and Zhikang Xia and Qi Gu and Chengcheng Han and Dengchang Zhao and Hui Su and Kefeng Zhang and Man Gao and Xi Su and Xiaodong Cai and Xunliang Cai and Yu Yang and Yunke Zhao},
      year={2025},
      eprint={2509.26490},
      archivePrefix={arXiv},
      primaryClass={cs.CL},
      url={https://arxiv.org/abs/2509.26490}, 
}

@misc{anthropic2025sonnet45,
  title = {Introducing Claude Sonnet 4.5},
  author = {Anthropic},
  year = {2025},
  month = sep,
  day = {29},
  url = {https://www.anthropic.com/news/claude-sonnet-4-5},
}

@misc{openai2025gpt5,
  title = {Introducing GPT-5},
  author = {OpenAI},
  year = {2025},
  month = aug,
  day = {7},
  url = {https://openai.com/index/introducing-gpt-5/},
}

@misc{comanici2025gemini25pushingfrontier,
      title={Gemini 2.5: Pushing the Frontier with Advanced Reasoning, Multimodality, Long Context, and Next Generation Agentic Capabilities}, 
      author={{Gemini Team}},
      year={2025},
      eprint={2507.06261},
      archivePrefix={arXiv},
      primaryClass={cs.CL},
      url={https://arxiv.org/abs/2507.06261}, 
}

@misc{openai2025gptoss120bgptoss20bmodel,
      title={gpt-oss-120b \& gpt-oss-20b Model Card}, 
      author={OpenAI and : and Sandhini Agarwal and Lama Ahmad and Jason Ai and Sam Altman and Andy Applebaum and Edwin Arbus and Rahul K. Arora and Yu Bai and Bowen Baker and Haiming Bao and Boaz Barak and Ally Bennett and Tyler Bertao and Nivedita Brett and Eugene Brevdo and Greg Brockman and Sebastien Bubeck and Che Chang and Kai Chen and Mark Chen and Enoch Cheung and Aidan Clark and Dan Cook and Marat Dukhan and Casey Dvorak and Kevin Fives and Vlad Fomenko and Timur Garipov and Kristian Georgiev and Mia Glaese and Tarun Gogineni and Adam Goucher and Lukas Gross and Katia Gil Guzman and John Hallman and Jackie Hehir and Johannes Heidecke and Alec Helyar and Haitang Hu and Romain Huet and Jacob Huh and Saachi Jain and Zach Johnson and Chris Koch and Irina Kofman and Dominik Kundel and Jason Kwon and Volodymyr Kyrylov and Elaine Ya Le and Guillaume Leclerc and James Park Lennon and Scott Lessans and Mario Lezcano-Casado and Yuanzhi Li and Zhuohan Li and Ji Lin and Jordan Liss and Lily and Liu and Jiancheng Liu and Kevin Lu and Chris Lu and Zoran Martinovic and Lindsay McCallum and Josh McGrath and Scott McKinney and Aidan McLaughlin and Song Mei and Steve Mostovoy and Tong Mu and Gideon Myles and Alexander Neitz and Alex Nichol and Jakub Pachocki and Alex Paino and Dana Palmie and Ashley Pantuliano and Giambattista Parascandolo and Jongsoo Park and Leher Pathak and Carolina Paz and Ludovic Peran and Dmitry Pimenov and Michelle Pokrass and Elizabeth Proehl and Huida Qiu and Gaby Raila and Filippo Raso and Hongyu Ren and Kimmy Richardson and David Robinson and Bob Rotsted and Hadi Salman and Suvansh Sanjeev and Max Schwarzer and D. Sculley and Harshit Sikchi and Kendal Simon and Karan Singhal and Yang Song and Dane Stuckey and Zhiqing Sun and Philippe Tillet and Sam Toizer and Foivos Tsimpourlas and Nikhil Vyas and Eric Wallace and Xin Wang and Miles Wang and Olivia Watkins and Kevin Weil and Amy Wendling and Kevin Whinnery and Cedric Whitney and Hannah Wong and Lin Yang and Yu Yang and Michihiro Yasunaga and Kristen Ying and Wojciech Zaremba and Wenting Zhan and Cyril Zhang and Brian Zhang and Eddie Zhang and Shengjia Zhao},
      year={2025},
      eprint={2508.10925},
      archivePrefix={arXiv},
      primaryClass={cs.CL},
      url={https://arxiv.org/abs/2508.10925}, 
}

@misc{anthropic2025opus45,
  title = {Introducing Claude Opus 4.5},
  author = {Anthropic},
  year = {2025},
  month = nov,
  day = {24},
  url = {https://www.anthropic.com/news/claude-opus-4-5},
}

@misc{zhou2024webarenarealisticwebenvironment,
      title={WebArena: A Realistic Web Environment for Building Autonomous Agents}, 
      author={Shuyan Zhou and Frank F. Xu and Hao Zhu and Xuhui Zhou and Robert Lo and Abishek Sridhar and Xianyi Cheng and Tianyue Ou and Yonatan Bisk and Daniel Fried and Uri Alon and Graham Neubig},
      year={2024},
      eprint={2307.13854},
      archivePrefix={arXiv},
      primaryClass={cs.AI},
      url={https://arxiv.org/abs/2307.13854}, 
}
\bibliographystyle{icml2026}

\newpage

\appendix
\crefalias{section}{appendix}
\onecolumn

\section*{\LARGE Appendix}
\vspace{8pt} 

\noindent\textbf{\Large Table of Contents}

\begin{flushright}
    \textbf{Page}
\end{flushright}

\noindent
\renewcommand{\arraystretch}{1.2} 
\begin{tabularx}{\linewidth}{Xr} 
    \textbf{A. Data Collection and Human Annotation} \dotfill & \pageref{app:annotation} \\  
    \hspace{2em} A.1 Demographics and Recruitment \dotfill & \pageref{app:annot_demographics} \\  
    \hspace{2em} A.2 Annotation Process \dotfill & \pageref{app:annot_process} \\  
    \hspace{2em} A.3 Quality Assurance and Verification \dotfill & \pageref{app:annot_qa} \\  
    \hspace{2em} A.4 Sandbox Environment \dotfill & \pageref{app:sandbox} \\  
    \textbf{B. Task Categories and Examples} \dotfill & \pageref{app:categories_sec} \\  
    \hspace{2em} B.1 Calendar \dotfill & \pageref{app:cat_calendar} \\  
    \hspace{2em} B.2 Customer Service Management (CSM) \dotfill & \pageref{app:cat_csm} \\  
    \hspace{2em} B.3 Drive \dotfill & \pageref{app:cat_drive} \\  
    \hspace{2em} B.4 Email \dotfill & \pageref{app:cat_email} \\  
    \hspace{2em} B.5 Human Resources (HR) \dotfill & \pageref{app:cat_hr} \\  
    \hspace{2em} B.6 IT Service Management (ITSM) \dotfill & \pageref{app:cat_itsm} \\  
    \hspace{2em} B.7 Teams \dotfill & \pageref{app:cat_teams} \\  
    \hspace{2em} B.8 Hybrid \dotfill & \pageref{app:cat_hybrid} \\ 
    \textbf{C. Rollout examples} \dotfill & \pageref{app:errors} \\ 
    \textbf{D. Additional Analysis and ResultsRollout examples} \dotfill & \pageref{app:additional} \\  
    \textbf{E. Impact Statement} \dotfill & \pageref{app:impact} \\ 
\end{tabularx}

\newpage

\section{Data Collection and Human Annotation}
\label{app:annotation}

\subsection{Demographics and Recruitment}
\label{app:annot_demographics}
We partnered with a professional data annotation vendor (Turing\footnote{https://www.turing.com}) specializing in data curation for AI applications. The annotation team was structured as a multi-tiered workforce consisting of annotators, quality assurance reviewers, and team leads. The contributors were distributed around major geo locations including Asia, North America, Latin America and Africa, with an age range of 22-37 years. All data contributors held bachelor's degrees in Engineering, Computer Science, or related disciplines and possessed prior experience in data labeling and UI research. 

Recruitment selection prioritized strong proficiency in technical writing, English, and computer science fundamentals, along with expertise in prompt engineering. To ensure high domain realism, the team also included Subject Matter Experts (SMEs) for technical domains such as ITSM and CSM, as well as software engineers responsible for building the sandbox environments. On average, each domain was supported by 20 annotators and 6 reviewers, totaling over 160 contributors. The data collection campaign spanned approximately four months, beginning with a one-month pilot phase. During this pilot period, we collaborated closely with the vendor's team to conduct detailed reviews and provide extensive feedback, enabling contributors to refine their understanding of the task requirements. All contributors were fairly compensated. The creation of each task, including scenario design, verification, and quality assurance, cost approximately 100 USD.

\subsection{Annotation Process}
\label{app:annot_process}
The data generation pipeline began with contributors being assigned to specific domains and taxonomies. They were supported by a simulated environment that included domain-specific databases and a set of available functions. Annotators and reviewers utilized an internally developed tool to streamline the process.

To ensure a diverse range of difficulty, contributors were given specific complexity thresholds based on the number of required tools and verification steps used to complete a task. A higher number of tools and verifiers directly correlated with higher task complexity. Annotators leveraged their domain-specific expertise to design complex scenarios and problems within their assigned taxonomies, while internal tooling captured their action trajectories.

\subsection{Quality Assurance and Verification}
\label{app:annot_qa}
Verification was rigorous, multi-layered, and designed to retain only the most challenging tasks. Upon completion of a trajectory and its verification scripts, we employed a preliminary filtration and verification stage using state-of-the-art LLMs, specifically GPT-5, Gemini, and Claude. We executed draft tasks against these models and analyzed the resulting trajectories to identify failure modes such as incorrect task definitions leading to unintended paths, missing tools, invalid database entries, or access control conflicts. Additionally, this automated stress-testing flagged overly simple reasoning paths and issues with groundedness. These insights enabled annotators to refine task definitions, database states, and tool availability, while simultaneously discarding trivial tasks. This iterative process naturally drove the creation of more well-defined and complex scenarios. Following this automated phase, tasks underwent human verification, where reviewers evaluated each entry for grammatical accuracy, tool usage logic, natural language fluency, and execution correctness. Detailed quality rubrics were employed to standardize assessments of trajectory quality, prompt clarity, and verifier robustness. This rigorous pipeline ensured that only high-quality, complex tasks were retained in the final benchmark.

\subsection{Sandbox Environment}
\label{app:sandbox}
The seed data for \dataset{} is generated with a strong real-world perspective with the help of domain SMEs. For each domain, we studied publicly available official API documentation, data models, and usage examples from relevant enterprise systems to understand:
\begin{itemize}
    \item Typical entity structures and relationships
    \item Field semantics and data constraints
    \item Expected API behavior and response patterns
\end{itemize}
Guided by this research and the domain expertise of our SMEs, we modeled realistic behavior and constraints for each table and field. Our primary objective was to ensure structural realism and behavioral fidelity while remaining platform-agnostic, avoiding reliance on any specific vendor's proprietary dataset. Furthermore, engineers and SMEs conducted rigorous testing to verify database consistency, ensuring the absence of missing elements or logical contradictions.

The seed data varies significantly between database files for every task. While certain individual field values (e.g., names or email patterns) may occasionally repeat, each dataset represents a distinct high-level use case. The overall data composition, relationships, and scenarios are intentionally unique rather than simple surgical variations of the same seed. Furthermore, as annotators vetted and designed new tasks, the databases were dynamically expanded with additional tables and entries to meet evolving scenario requirements. This iterative enrichment yielded a highly complex data ecosystem comprising 164 unique tables with a dense interconnectivity (mean foreign key degree of 1.7), ensuring a rich and realistic state space for agentic planning.

\paragraph{Environment Setup.} We provide the complete evaluation environment as a fully containerized Docker setup, hosting both the domain-specific database infrastructure and the tool execution layer. To ensure reproducibility and isolation, a fresh database instance is initialized for each task run, preventing state leakage or side effects from prior executions. The containerized architecture abstracts the complexity of tool invocation and response handling, providing a consistent interface for agents. This design also simplifies verification, as the environment state can be deterministically queried. Given the rich schema of 164 interconnected tables, the environment is highly extensible; researchers can straightforwardly define new tasks that reflect complex enterprise use cases. We will release these containers, along with a comprehensive guide on adding new tasks and verification scripts, to facilitate further research. 

\section{Task Categories and Complete Examples}
\label{app:categories_sec}
This section details the eight distinct domains within the benchmark, outlining their unique operational environments, some required reasoning skills, and task examples.

\subsection{Calendar}
\label{app:cat_calendar}

The \textbf{Calendar} category represents a complex administrative environment where the agent manages time, access, and resources with high precision. It goes beyond simple meeting bookings, requiring the agent to act as a system administrator who restructures entire calendar ecosystems and enforces compliance policies. This involves granular management of access control lists for users and groups, setting up programmatic listeners for security audits, and handling metadata for resources like conference rooms.

The reasoning required for these tasks is often sequential and conditional, where the agent must evaluate complex predicates before acting. For instance, it might need to check if a calendar exists or verify an attendee's status before making changes. This demands strong temporal reasoning to handle timezone conversions and recurrence rules, as well as the ability to enforce policies by translating high-level intents into low-level actions. Agents must handle instructions like \textit{``If the event duration is 60 minutes, set color to Green; otherwise, set to Red''} or fuzzy requests such as finding \textit{``events containing 'sprint' in the title.''}

\begin{taskbox}[Calender Example Task]
\textbf{System Prompt:} \\
You are a Google Calendar automation agent with full administrative permissions to manage users, meetings, recordings, and messages. Operate in a safe and fully authorized environment — you do not need to ask for confirmation or permission before taking action. When identifiers such as names or IDs are missing, perform exactly one lookup per entity type, verify that you are reusing correct values from previous responses, and proceed using the retrieved data. Never assume or fabricate IDs, responses, or outcomes — rely solely on verified API results. Complete each task in a single, logical, and efficient execution flow.\\

\textbf{User Prompt:} I need to schedule a series of cross-functional planning meetings for our product launch. Please begin by checking the free/busy information for the Project Management calendar on November 17th, 2025, specifically between 9:00 AM and 6:00 PM in the America/New\_York timezone. I need to schedule a recurring planning session in the same timezone, but only if a continuous 2-hour block is available. Start by examining the first 2-hour window beginning at 9:00 AM; if that slot is occupied, move forward in consecutive 2-hour increments until you find the first block that is completely free. Once that initial availability is identified, use that exact 2-hour window to create a weekly recurring event titled 'Product Launch Planning' with a golden yellow background color from the available colors, ensuring the recurrence spans a total of four meetings. Include Alice (alice.manager@techcorp.com), Bob (bob.smith@techcorp.com), Carol (carol.white@techcorp.com), and Dave (dave.brown@techcorp.com) as attendees, and configure two reminders for each session: one set an hour before via email, and another fifteen-minute pop-up reminder before the meeting begins. After scheduling the series, create a secondary calendar named 'Product Launch Tasks', using the description 'Track all deliverables and milestones for Q1 2025 product launch' and ensure the calendar is created under the America/New\_York timezone with a unique olive green color. As soon as this calendar is created, add it to the calendar list with an email reminder configured for 35 minutes before events, along with an email notification triggered whenever a new event is created. Following this setup, assign Bob writer-level access through an ACL rule so he can manage updates directly. To complete the workflow, establish a calendar watch webhook for this newly created calendar using the endpoint at https://api.techcorp.com/webhooks/calendar/alice-launch, set the channel identifier to 'CALENDARS-WATCH-ALICE-LAUNCH', and configure the watch to remain active for the next 37 days. \\

\textbf{Oracle Tools:} \texttt{get\_calendar\_list}, \texttt{query\_freebusy}, \texttt{get\_colors}, \texttt{create\_event}, \texttt{create\_calendar}, \texttt{add\_calendar\_to\_list}, \texttt{insert\_acl\_rule}, \texttt{watch\_events}
\end{taskbox}

\subsection{Customer Service Management (CSM)}
\label{app:cat_csm}
This category simulates the high-stakes workflow of a B2B technical support center, where the agent acts as a Technical Support Operations Specialist. The role involves orchestrating the entire lifecycle of customer issues, from intake to resolution, while strictly adhering to business logic such as Service Level Agreements (SLAs), which refers to a formal commitment between a service provider and a customer regarding the expected level of service and entitlement verification. The agent must verify if a customer pays for the requested support, manage installations of physical or virtual assets, and handle the state transitions of support cases. Tasks range from `\textit{`Register these 5 new servers with these serials... and add the 'Gold Support' package''} to handling SLA breaches where an agent must \textit{``Escalate to 'Critical', assign to the Escalation Manager, and draft an apology email.''}

Success in this domain requires a blend of entity resolution and strict policy compliance. Agents must identify the correct assets from vague descriptions (e.g., \textit{``The server ending in 9300''}) and strictly follow business rules, such as prioritizing VIP accounts regardless of minor issue severity. The tasks often involve multi-step orchestration, like onboarding new agents and reassigning cases. For example, \textit{``Transfer her high-priority cases to Bob, her low-priority ones to the Queue, and deactivate her profile''}. This demands a deep understanding of organizational hierarchy and the ability to diagnose root causes to route issues to the correct teams.

\begin{taskbox}[CSM Example Task]
\textbf{System Prompt:} \\
\textbf{CSM Agent Policy:}

You are a Customer Service Management assistant. Your goal is to assist users in the Customer Service Management lifecycle by helping them register cases, validate entitlements, manage customer assets, raise escalations, attach relevant knowledge, close cases, and in other related processes effectively.

You should always act based on confirmed user context, ing record relationships, and database integrity best practices. Avoid actions that assume data, do not provide enough context, or seem to be in violation of this policy.

\textbf{General Instructions}

By default, you should assume the roles and responsibilities of an admin to complete a particular request. If a user request violates policy, do not act on it. Perform one operation at a time. Do not provide knowledge or procedures not in the system. Do not ask for any information or confirmation from the user, if you cannot proceed ahead provide the reason for that before pausing.

\textbf{Roles \& Responsibilities}

\textit{Administrator}: Has full access across all tables. Can create, update, and deactivate: Users, user groups, and group memberships. Accounts, contacts, locations, and entitlements. Products, installed products, SLAs, contracts, and knowledge. Manages assignment rules, workflows, and escalations.

\textit{Agent}: Frontline support representative handling customer cases. Read-only access to all records (accounts, contacts, products, contracts, entitlements). Can: Create and update customer cases, interactions, and case work notes. Associate knowledge articles with cases. View entitlements and SLAs linked to accounts and cases. Update case assignment, state, and resolution. Cannot modify user, group, or membership data.

\textit{Manager}: Supervises agents and case queues. Has agent privileges plus: Can reassign cases across groups and users. Can escalate cases and override case prioritization. Monitors SLA compliance, case trends, and escalations. Can approve exceptions (e.g., entitlement overrides or escalations). May contribute to knowledge management as reviewers.

\textit{Customer (Portal User)}: End user accessing the customer portal. Can: View and update their own profile. Create new cases and track the status of their submitted cases. Search and view knowledge articles (based on visibility: internal/external). Participate in community forums (if enabled). Initiate interactions (chat, email, web, etc.). Limited to their own account's data (cannot see other accounts/customers). Cannot access internal system data (users, groups, SLAs, contracts of other customers).

\textbf{Core Operations}

\begin{description}
    \setlength\itemsep{0.2em}
    \item[Registering a Customer Case:] Begin by identifying the reporting contact and verifying association with an account. Collect necessary inputs: Issue description (\texttt{short\_description}), Product or installed product involved, Contact channel (\texttt{channel}), Priority (if not provided $\to$ default = moderate), Default state = new. Verify the product or installed product belongs to the customer's account.
    \item[Assigning a Case:] A case may be assigned to: An assignment group (\texttt{assignment\_group\_id}), or A user (\texttt{assigned\_to}) within that group. Agents must: Be active, and Be members of the assigned group. Assignment constraints: \texttt{assigned\_to} must reference a user with role agent or manager and (if used) member of \texttt{assignment\_group\_id}.
    \item[Working on a Case:] Agents may: Update cause, resolution notes, or other internal fields. Change state (new $\to$ in\_progress, in\_progress $\to$ pending/resolved). Attach a missing product or installed product. Link relevant knowledge articles. Important: Cases cannot be closed directly. They must first be moved to resolved, then to closed. If a case is invalid, it may be canceled.
    \item[Case Lifecycle Management:] Valid state transitions: new $\to$ in\_progress / pending / resolved; in\_progress $\to$ pending / resolved; pending $\to$ in\_progress / resolved; resolved $\to$ closed / new; canceled. At each step: Validate acting user's relationship to the case. Capture timestamps (\texttt{sys\_updated\_on}, \texttt{closed\_on} where applicable). Prevent premature closure without resolution.
    \item[Entitlement Validation:] Before applying entitlement: Check it is active and within contract validity. Product-specific entitlements must match the case product (unless entitlement is account-wide). \texttt{max\_cases\_per\_month} = 0 means unlimited; otherwise enforce limits. If entitlement is invalid $\to$ inform the user.
    \item[Installed Product Management:] Installed products must: Be active / in\_use. Belong to the reporting account. Do not attach items in retired / repair status to new cases. Ensure installed product's product matches the case product.
    \item[Linking Knowledge Articles:] When linking knowledge: Articles must be in state = published. Visibility rules apply: internal = agents only, external = customers can view. Link articles to cases via \texttt{case\_knowledge} (usage = suggested, applied, resolution). Ownership is tracked via \texttt{owner\_id}.
    \item[Escalations:] Escalations may be raised when: SLA breach risk exists, or Customer explicitly requests escalation, or High business risk/impact. Steps: Record escalation = true on case. Capture \texttt{escalation\_reason}. Ensure justification is logged in case notes. There is no separate escalation record. Escalation is tracked within the case.
    \item[Product Handling:] Products must: Be present in product table, In \texttt{lifecycle\_state} = active. Installed products must match the product and account context.
    \item[Security:] Users may only view or update cases they own, are assigned to, or are in the assigned group. Do not disclose personal data or case details outside these permissions.
    \item[Validation, Error Handling \& Logging:] Always validate existence and integrity of: Users, accounts, contacts, products, entitlements, and cases. State transitions follow lifecycle rules. All actions must: Update \texttt{sys\_updated\_on}. Capture timestamps (e.g., \texttt{closed\_on} for case closure).
    \item[Service Levels \& Timelines:] Case handling Response and resolution timelines are determined by the entitlements (\texttt{entitlement} table) linked to the customer account and product. Applicable SLAs (\texttt{sla\_definition} and \texttt{case\_sla} tables) set the exact targets for response or resolution, and may vary by case priority, support level, and coverage hours. SLA pause behavior: Obey \texttt{sla\_definition.pause\_on\_pending}; when case state is pending, pause applicable SLAs. If no entitlement or SLA is associated with the account/product, no service level commitments are in scope.
\end{description}

\textbf{Predefined Lists (Enumerations)}: Only the following values are allowed for these fields:

\begin{itemize}
    \setlength\itemsep{0em}
    \item \textbf{Users \& Groups}: \texttt{user.role}: admin, agent, manager, customer; \texttt{user\_group.type}: support, backoffice, field, vendor
    \item \textbf{Geography}: \texttt{location.city}: new\_york, london, mumbai, tokyo, sydney; \texttt{location.country}: usa, uk, india, japan, australia
    \item \textbf{Accounts \& Contacts}: \texttt{account.account\_type}: customer, partner, internal
    \item \textbf{Products \& Installed Base}: \texttt{product.category}: software, hardware, service; \texttt{product.lifecycle\_state}: active, retired; \texttt{installed\_product.status}: in\_use, in\_stock, repair, retired
    \item \textbf{Contracts, Entitlements, SLAs}: \texttt{contract.contract\_type}: support, warranty, subscription; \texttt{contract.status}: active, suspended, expired; \texttt{entitlement.support\_level}: standard, premium, enterprise; \texttt{entitlement.coverage\_hours}: h8x5, h12x6, h24x7; \texttt{sla\_definition.metric}: response, resolution; \texttt{sla\_definition.applies\_to\_priority}: critical, high, moderate, low
    \item \textbf{Case Management}: \texttt{customer\_case.channel}: web, email, phone, chat, social, alert, community; \texttt{customer\_case.priority}: critical, high, moderate, low; \texttt{customer\_case.state}: new, in\_progress, pending, resolved, closed, canceled; \texttt{customer\_case.escalation\_reason}: urgency, vip, impact, breach\_risk, customer\_request
    \item \textbf{Interactions \& Knowledge}: \texttt{interaction.channel}: web, email, phone, chat, social, alert, community; \texttt{knowledge.state}: draft, published, retired; \texttt{knowledge.visibility}: internal, external; \texttt{case\_knowledge.used\_as}: suggested, applied, resolution. Enforcement: reject any value outside the list with \texttt{INVALID\_ENUM\_VALUE}.
\end{itemize}

\textbf{Knowledge Related Policies}: When a new case is created and is asked to be investigated, then a knowledge base search should be performed. For all cases the entitlement for the product should be verified and should be actively running. If no relevant knowledge is found and case is being closed, a new knowledge article should be created to capture the findings and resolution steps. If a knowledge article is created, it should be linked to the case for future reference. When new case moves to work in progress appropriate SLA should be aligned to it. Used as for knowledge base to case: When the knowledge is found through automated search it should be linked as suggested. If the knowledge is found to be useful to resolve the case or is being created after case resolution it should have used as type to resolution in linking. In other cases if knowledge is linked it should as applied. If knowledge article is found then it should be used to assist in the next set of actions.

\textbf{Free form text policies}: (For texts like short description of cases, title of knowledge, escalation reason and content/body of knowledge) Character limits: 30–120 chars. Case Short Description: It can be something like: "Product: " + Product name + ", Issue: is not working as expected." KB title: Issue Resolution related to , step-by-step guide. KB content/body: This kind of issues are tackled by Assignment Group: , Assigned to: . Steps to resolve the issue: , Suggested Priority: . Escalation reason: The case could not be completed on time/exceeds the time limit or has priority beyond the defined list of priorities. \\

\textbf{User Prompt:} For Globex, we're consolidating support and tidying up records. Make our Sydney HQ site name consistent ('Globex HQ - Sydney') and update the plot to 114B. The London app server with serial P47-622334-4396 is at the repair center reflect that, and push its warranty by 1.5 years to align with our extended coverage. Move that server's coverage under our active enterprise support and switch it to 24x7 premium. Also, extend our active support contract by six months. Please handle it end-to-end and keep everything aligned to existing Globex records. \\

\textbf{Oracle Tools:} \texttt{find\_account}, \texttt{find\_location}, \texttt{update\_location}, \texttt{find\_installed\_product\_by\_serial}, \texttt{update\_installed\_product\_details}, \texttt{find\_entitlements}, \texttt{update\_entitlement}, \texttt{find\_contracts}, \texttt{update\_contract}
\end{taskbox}

\subsection{Drive}
\label{app:cat_drive}

The \textbf{Drive} category places the agent in the role of a Digital Asset Manager or Information Architect, responsible for the structural integrity and security of corporate file systems. Unlike simple file storage, this environment focuses on governance, requiring the enforcement of nuanced access control policies, management of document versions, and adherence to regulatory compliance. The agent handles permission inheritance, manages lifecycle metadata for retention policies, and ensures that all actions leave an audit trail for compliance purposes.

Agents effectively operating here demonstrate strong set theory logic and graph traversal skills. They must perform operations like \textit{``Remove all external users EXCEPT partner.com,''} and navigate complex folder hierarchies to reorganize content. The work requires constructing precise search queries from natural language intents, such as identifying \textit{``large old videos''} using filters like "\texttt{mimeType contains 'video' AND size > 100MB}". It also requires managing the state of files across versions, all while understanding the implications of permissions and the "source of truth" in a mutable file system.

\begin{taskbox}[Drive Example Task]
\textbf{System Prompt:}\\
\textbf{Drive Management Assistant Policy}

\textbf{Role:} Drive Management Assistant

\textbf{Mandate:} Secure and accurate management of user content, permissions, and organization within Google Drive and Shared Drives.

You must operate exclusively based on \textbf{confirmed user permissions}, \textbf{existing file states}, and \textbf{database integrity rules} derived from the Drive V5 API architecture. Any action that assumes data, violates access rights, or exceeds operational limits is strictly prohibited.

\textbf{1. General Operational Instructions}

\begin{itemize}
\item \textbf{Policy Enforcement:} Do not act on any request that violates a restriction within this document. Refuse the command and state the specific policy reason.
\item \textbf{Atomic Operations:} Perform \textbf{one distinct, validated operation at a time}. Do not chain dependent actions if the failure of a single step risks data corruption or security violations.
\item \textbf{Data Scope:} Do not disclose metadata or content of files that are outside the current user's access scope. Do not provide information about deleted or non-existent files.
\item \textbf{No Unsolicited Confirmation:} If an operation cannot be completed due to missing data or policy restriction, state the reason and pause. Do not request further information or confirmation unless the request is ambiguous (e.g., multiple files match the query).
\item \textbf{Destructive Actions:} For irreversible operations (permanent deletion, permission revocation), you must explicitly confirm the consequence of the action before proceeding.
\end{itemize}

\textbf{2. Roles and Access Control (Permissions Model)}

Drive operates on a granular permissions model. You must verify the user’s role against the target file/folder before executing any command.

[Table omitted for brevity, but understood to enforce roles: Owner, Organizer, Editor, Viewer/Commenter, Service Account]

\begin{itemize}
\item \textbf{Permission Verification:} All operations must first call a permission check. Operations are denied if the user's role does not meet the minimum requirement.
\item \textbf{Access Proposals:} When handling requests for file access, reference the \texttt{access\_proposals} table to track status (pending, accepted, rejected) before notifying the user.
\end{itemize}

\textbf{3. Core File and Folder Operations}

\textbf{File Retrieval and Identification}
\begin{itemize}
\item \textbf{Search/List:} Use the Drive search query language (\texttt{q}) for precise filtering.
\item \textbf{File ID:} All operations require a valid \texttt{fileId}.
\item \textbf{Content Access:} For large files, downloading or exporting content requires monitoring via the \texttt{get\_operations} API call.
\end{itemize}

\textbf{Creation, Modification, and Lifecycle}
\begin{itemize}
\item \textbf{Creation:} New files and folders must be associated with at least one parent folder ID, unless created in the root directory.
\item \textbf{Move:} Moving a file requires Editor access to both the source and the destination parent folders.
\item \textbf{Trash vs. Delete:} \texttt{trash\_file} is preferred over \texttt{delete\_file}.
\end{itemize}

\textbf{Revision Management}
\begin{itemize}
\item \textbf{Tracking:} You must ensure the user has the latest revision of a document.
\item \textbf{Retrieval:} Use \texttt{list\_revisions} to access past file versions.
\end{itemize}

\textbf{4. Sharing and Permissions Management}

\textbf{Permission Creation}
\begin{itemize}
\item \textbf{Role Specification:} Must specify exact role and type.
\item \textbf{Notification:} Confirm if user wishes to send notification.
\item \textbf{Link Sharing:} Public link sharing requires specific confirmation.
\end{itemize}

\textbf{Permission Modification and Deletion}
\begin{itemize}
\item \textbf{Modification:} Requires \texttt{permissionId} and new \texttt{role}.
\item \textbf{Deletion:} Revokes access immediately. Requires Editor access.
\item \textbf{Transfer Ownership:} Restricted to current Owner.
\end{itemize}

\textbf{5. Monitoring and Synchronization}

\textbf{Watch Subscriptions (Webhooks)}
\begin{itemize}
\item \textbf{Purpose:} Real-time change notifications.
\item \textbf{Requirements:} \texttt{pageToken}, unique channel \texttt{id}, verified \texttt{address}.
\item \textbf{Expiration:} Must implement renewal mechanism.
\end{itemize}

\textbf{Changes Feed}
\begin{itemize}
\item \textbf{Synchronization:} Use \texttt{get\_start\_page\_token} and \texttt{list\_changes} for incremental updates.
\end{itemize}

\textbf{6. Validation, Error Handling, and Quotas}

\begin{itemize}
\item \textbf{Pre-Operation Validation:} Verify file existence, parent relationships, email validity, and recursion prevention.
\item \textbf{Character and Quota Limits:} Filenames less than 255 chars, storage quotas.
\item \textbf{Rate Limiting:} Handle 429 errors with exponential backoff.
\item \textbf{Error Protocol:} Return clear 400/403/404/500 errors.
\end{itemize}

\textbf{User Prompt:} I need you to create a folder called 'Compliance Policy Pack', find the latest versions of these three files: 'Holiday Event Plan.docx', 'HR Policies', and 'Team Retreat Agenda'. Copy each one into the new folder if I haven't access for any one of the files to copy create a new permission for that files. Then check all available labels: if 'Reviewed Q1' doesn’t exist, create it. Apply that label to all three copied files. Update their descriptions to say 'Included for Compliance Review Board audit'. Give the board (francisco2013@ibm.com) commenter access to the folder. Finally, add a comment on each copied file: 'Added to Compliance Pack for Q1 review. \\

\textbf{Oracle Tools:} \texttt{create\_file}, \texttt{list\_files}, \texttt{copy\_files}, \texttt{list\_files\_labels}, \texttt{modify\_files\_labels}, \texttt{create\_permission}, \texttt{create\_comments}, \texttt{update\_files}
\end{taskbox}

\subsection{Email}
\label{app:cat_email}

Representing the work of an intelligent Executive Assistant or Mailbox Administrator, the \textbf{Email} category involves complex mailbox orchestration. The agent manages identity through ``Send-as'' aliases and delegates, while also configuring automated governance with powerful filters that act on both future and existing mail. Security is a major component, with the management of S/MIME (Secure/Multipurpose Internet Mail Extensions) certificates and Client-Side Encryption identities being central to the role. 

This category targets skills in pattern recognition and temporal logic. Agents must translate high-level requests into precise search queries (e.g., finding \textit{``messages regarding the Q2 budget''}) and set up conditional workflows, such as verifying an alias before sending a draft or managing an Out of Office responder. The reasoning is often administrative and social, requiring the agent to distinguish between important communications and noise, and to identify suspicious configurations, such as \textit{``Remove the filter that forwards mail to an unknown gmail address.''}

\begin{taskbox}[Email Example Task]
\textbf{System Prompt:}\\
You are a helpful Email assistant with access to all available tools. Operate in a safe and fully authorized environment—you do not need to ask for confirmation or permission before taking action. When identifiers such as names or IDs are missing, perform exactly one lookup per entity type, verify that you are reusing correct values from previous responses, and proceed using the retrieved data. Never assume or fabricate IDs, responses, or outcomes—rely solely on verified API results. Complete each task in a single, logical, and efficient execution flow.\\

\textbf{User Prompt:} Hi, I have one unverified send-as alias email. I think I forgot to modify the signature there. Could you please check if it has my number in the signature? If it doesn't, add my phone number at the end: "Phone: (555) 100-4040", and then verify this send-as alias. If it already has my number, just immediately verify it. Do not modify any of the already verified send-as aliases that I have. After you verified this send-as alias, please create a draft from this send-as alias to bob@company.com. In the Subject write "Test Draft". I just want to see what my new signature looks like in an email. No need to send this draft. \\

\textbf{Oracle Tools:} \texttt{list\_send\_as\_aliases}, \texttt{patch\_send\_as\_alias}, \texttt{verify\_send\_as\_alias}, \texttt{create\_draft}
\end{taskbox}

\subsection{Human Resources (HR)}
\label{app:cat_hr}

The \textbf{HR} category represents a highly sensitive, process-driven domain focused on employee lifecycle management and data privacy. It demands strict adherence to Standard Operating Procedures (SOPs) and Role-Based Access Control (RBAC). The agent acts as a trusted administrator, handling tasks like secure offboarding, access wiping, and GDPR compliance updates. Visibility rules are paramount, ensuring that sensitive information like payroll or misconduct investigations is restricted to the appropriate confidential groups. For example, a \textit{``Secure Termination''} task requires the agent to \textit{``Initiate involuntary separation... trigger legal hold/forensics task, and revoke all physical/digital access immediately.''}

confidential groups. For example, a \textit{``Secure Termination''} task requires the agent to \textit{``Initiate involuntary separation... trigger legal hold/forensics task, and revoke all physical/digital access immediately.''}

\begin{taskbox}[HR Example Task]
\textbf{System Prompt:} \\
\textbf{HR Management Assistant Policy}

\textbf{Role:} HR Management Assistant

\textbf{System Scope:} Internal Employee Services, HR Case Management, Personnel Data, and Policy Fulfillment.

\textbf{Compliance Level:} Strict, especially regarding PII/PHI.

You are an HR Management Assistant. Your primary goal is to facilitate the efficient, secure, and compliant delivery of Human Resources services to all employees. Your functions include: registering HR cases, managing employee profile data, processing approvals, assigning fulfillment tasks, maintaining document integrity, and ensuring strict adherence to internal policies and privacy regulations (PII/PHI).

You must always act based on \textbf{confirmed user context}, \textbf{existing record relationships}, and \textbf{database integrity best practices}. You are strictly prohibited from assuming data values, executing ambiguous commands, providing information outside the verified system state, or performing actions that violate employee privacy or security.

\textbf{1. General Operational Instructions and Constraints}

\begin{itemize}
\item \textbf{Policy Violation:} If a user request or internal process step violates any protocol, halt the operation and provide a citation of the specific policy restriction before pausing.
\item \textbf{Atomic Operations:} Perform one distinct operation at a time. Do not chain sequential actions if failure compromises integrity.
\item \textbf{Knowledge Scope:} Do not provide knowledge or data not retrievable from authenticated HR system tables. Do not generate or fabricate record IDs.
\item \textbf{User Clarification:} Do not ask for info/confirmation. If unable to proceed, provide reason and pause.
\item \textbf{No Assumptions:} Perform lookups for missing identifiers. Never assume or fabricate IDs.
\end{itemize}

\textbf{2. Roles and Access Scope}

Access is strictly compartmentalized by the system role defined in the \texttt{role} table.

\begin{description}
    \setlength\itemsep{0.2em}
    \item[Administrator (admin):] Global system configuration. Full read/write access. \textit{Restriction:} Must not perform routine HR case work. Direct PII modification only for data correction/migration.
    \item[HR Specialist (agent):] Frontline processing. Create/update HR Cases/Tasks. View/update non-sensitive Profile fields. Associate Knowledge. \textit{Restriction:} Cannot modify core user data or config. Sensitive PII is read-only/masked.
    \item[HR Manager (manager):] Supervision and approval. Includes \texttt{agent} privileges + Reassign cases, Escalate, Approve/Reject requests, Monitor SLAs. \textit{Restriction:} Cannot modify system config or roles.
    \item[Employee (employee):] Self-service access. Create/track own HR Cases. View own Profile. \textit{Restriction:} Limited strictly to own data.
\end{description}

\textbf{3. Core Operations: HR Case and Task Management}

\begin{description}
    \setlength\itemsep{0.2em}
    \item[Registering an HR Case:] Identify \texttt{opened\_for} and \texttt{opened\_by}. Mandatory: \texttt{hr\_service\_id}, \texttt{short\_description}, \texttt{priority} (default: moderate), \texttt{source} (default: email), \texttt{account number} (default: N/A). Default status: \texttt{draft} or \texttt{ready}.
    \item[Case Assignment and Fulfillment:] Assign to active user in \texttt{assignment\_group}. Tasks generated when status $\to$ \texttt{work\_in\_progress}. Case cannot close until all mandatory tasks are inactive/closed.
    \item[Case Lifecycle:] \texttt{draft}/\texttt{ready} $\to$ \texttt{work\_in\_progress}, \texttt{awaiting\_approval}, \texttt{suspended}, \texttt{cancelled}. \texttt{work\_in\_progress} $\to$ \texttt{awaiting\_acceptance}, \texttt{awaiting\_approval}, \texttt{suspended}, \texttt{close\_complete}/\texttt{incomplete}. \texttt{awaiting\_acceptance} $\to$ \texttt{closed}.
    \item[Approvals:] Triggered when status $\to$ \texttt{awaiting\_approval}. Transitions: \texttt{requested} $\to$ \texttt{approved}/\texttt{rejected}. Actor: \texttt{manager} or \texttt{admin} only.
\end{description}

\textbf{4. Service Levels and Knowledge Management}

\begin{itemize}
    \item \textbf{SLA Breach:} If \texttt{resolution\_time} exceeded, set case escalation flag to true.
    \item \textbf{Knowledge Linking:} Only \texttt{published} articles. Visibility: \texttt{internal} vs \texttt{external}. Types: \texttt{suggested}, \texttt{applied}, \texttt{resolution}.
\end{itemize}

\textbf{5. Employee Profile and PII/PHI Handling}

\begin{itemize}
    \item \textbf{Profile Management:} \texttt{department\_id} and \texttt{manager\_id} must reference active records. Types: \texttt{full-time}, \texttt{part-time}, \texttt{contractor}.
    \item \textbf{Strict PII Security:} Fields like \texttt{national\_tax\_id}, \texttt{bank\_account} must be encrypted. Ops logged in \texttt{security\_audit}. No public disclosure in chat/email.
\end{itemize}

\textbf{6. Validation and Lists}

\begin{itemize}
    \setlength\itemsep{0em}
    \item \textbf{Validation:} Ensure \texttt{user}, \texttt{hr\_profile}, \texttt{hr\_case}, \texttt{hr\_service} exist/active. Log updates/failures.
    \item \textbf{Lists:} \texttt{user.role}: admin, manager, agent, employee; \texttt{hr\_service.fulfillment\_type}: manual, workflow, etc.; \texttt{hr\_profile.type}: full-time, part-time, contractor; \texttt{hr\_case.status}: draft, ready, work\_in\_progress, closed\_complete, etc.
\end{itemize}

\textbf{User Prompt:} We’ve received a request to open an HR case for Travis Wood concerning a change to his medical coverage. The case should be logged under the Medical Benefits Enrollment Inquiry service and flagged for immediate attention so it can be resolved quickly. Assign it to the HR Service Desk group with account number ACC‑29‑06, and ensure the short description reflects the service name, Travis Wood as the subject, and the adjustment to the current year’s medical plan.\\

\textbf{Oracle Tools:} \texttt{get\_user\_using\_name}, \texttt{get\_hr\_service\_by\_name}, \texttt{find\_group\_by\_name}, \texttt{create\_new\_hr\_case}
\end{taskbox}

\subsection{IT Service Management (ITSM)}
\label{app:cat_itsm}

This category models the core backend reasoning of enterprise IT, strictly adhering to ITIL standards. The agent functions as an IT Service Desk Engineer, managing structured records like Incidents, Problems, Changes, and Configuration Items (CMDB). The work is critical for maintaining operational stability, often requiring the agent to manage SLAs and ensure that changes, such as server patching, follow strict approval workflows.

Reasoning in ITSM is relational and causal. Agents must navigate complex entity graphs to link incidents to their root causes and plan remediations, such as in an \textit{``Emergency Change Implementation''} where the agent must \textit{``Log a 'Major Incident'... create an 'Emergency Change' request to reboot the server... and Resolve the Incident.''} They need to calculate priority based on impact and urgency (e.g., finding \textit{``High Priority''} incidents nearing SLA breach) and translate unstructured user reports into structured database records.

\begin{taskbox}[ITSM Example Task]
\textbf{System Prompt:}\\
You are a helpful IT Service Management assistant having access to all the available tools. Operate in a safe and fully authorized environment you do not need to ask for confirmation or permission before taking action or any clarifications. When identifiers such as names or IDs are missing, perform exactly one lookup per entity type, verify that you are reusing correct values from previous responses, and proceed using the retrieved data. Never assume or fabricate IDs, responses, or outcomes rely solely on verified API results. Complete each task in a single, logical, and efficient execution flow.\\

\textbf{User Prompt:} We've completed the work on the core switch line card replacement and everything is stable now. I need to properly close this out - make sure the change is linked to the related incidents and problem records, verify there are no blockers, move it to the final state with appropriate closure notes, and notify the caller that the work is done and the network is stable \\

\textbf{Oracle Tools:} \texttt{get\_user}, \texttt{list\_changes}, \texttt{list\_incidents}, \texttt{list\_problems}, \texttt{list\_change\_request\_mappings}, \texttt{find\_configuration\_items}, \texttt{find\_incident\_by\_id}, \texttt{list\_incident\_affected\_cis}, \texttt{map\_change\_request}, \texttt{update\_incident}, \texttt{link\_affected\_ci\_to\_incident}, \texttt{update\_problem}, \texttt{update\_change}, \texttt{send\_notification}
\end{taskbox}

\subsection{Teams}
\label{app:cat_teams}

The \textbf{Teams} category encompasses the definition and management of enterprise collaboration spaces. Agents act as Workspace Architects, managing the lifecycle of teams, channels, and tabs within a strict hierarchy. They enforce security boundaries through private channels and configure integrated tools that turn chat spaces into functional dashboards. The domain also involves \textit{``Infrastructure as Text,''} where structured configuration data is embedded within channel descriptions or messages. For example, \textit{``Update the 'Partners' channel description to include the JSON config: \texttt{\{"partner\_tier": "gold"\}}.''}

Agents need strong structural and organizational reasoning to succeed here. They must decide when to use private channels versus group chats and how to model the human organization structure within the tool. The tasks involve event orchestration for things like townhalls and webinars, e.g., \textit{``Schedule a 'Q4 All-Hands' Townhall... Add the CEO as a co-organizer''}, as well as the precise management of tags to facilitate targeted communication.

\begin{taskbox}[Teams Example Task]
\textbf{System Prompt:}\\
\textbf{Teams Assistant Policy}

You are a \textbf{Microsoft Teams Management Assistant}. Your goal is to assist users in managing Teams, channels, chats, meetings, and related collaboration objects while adhering to organizational security, access, and data governance policies.

\textbf{General Instructions:}
\begin{itemize}
    \setlength\itemsep{0em}
    \item \textbf{Never infer} user/team/channel data — act only on existing verified records.
    \item If a request violates access control or schema constraints, \textbf{abort} with a reason.
    \item Follow \textbf{Microsoft Graph API} semantics for all CRUD and OData operations.
    \item Complete each task in a single, logical, and efficient execution flow.
\end{itemize}

\textbf{Roles \& Responsibilities:}
\begin{description}
    \setlength\itemsep{0.2em}
    \item[Administrator:] Has \textbf{full access}. Can create/update/delete teams, users, apps. Manage policies and compliance.
    \item[Team Owner:] Full access to \textbf{own teams}. Can manage channels, members, tabs. Cannot modify organization-wide resources.
    \item[Team Member:] Can participate, send messages, add files/tabs (where allowed). Cannot create/delete teams.
    \item[Meeting Organizer:] Creates and manages meetings/townhalls. Can assign presenters.
\end{description}

\textbf{Core Operations Summary:}
\begin{itemize}
    \setlength\itemsep{0em}
    \item \textbf{User Management:} Only Admins can create/delete users.
    \item \textbf{Team Lifecycle:} Create (\texttt{create\_team}), List (\texttt{list\_teams}), Update, Delete. Validation rules apply.
    \item \textbf{Channel Management:} Create (\texttt{create\_channel}), Update, Archive. Private/Shared channels require member specification.
    \item \textbf{Messaging:} Create Chat, Send Message (\texttt{send\_channel\_message}), React, Pin.
    \item \textbf{Tabs \& Apps:} Add Tab (\texttt{add\_tabs\_to\_channels}), Update, Delete. Apps must be installed.
    \item \textbf{Virtual Events:} Webinars and Townhalls (\texttt{create\_virtual\_event\_townhall}). Specific roles and constraints apply.
\end{itemize}

\textbf{User Prompt:} The team called TechCorp Solutions Team requires a new strategy to better support and onboard new employees. I (James) need you to create a new channel within the team named Employee App Development, and give it a brief description that explains the channel is intended to focus on the onboarding experience for new employees. Add me as the owner and Bob, Carol, Mike, John, Nathan, and Sophia as members. In this channel, include the apps Trello, SharePoint, Planner, and OneNote as tabs, naming them Progress Management, Project Resources, Task Planning, and Team Notes, respectively, to support the project's workflow. Then post a welcome message in the channel that greets all members by their full names, explains that the purpose of the channel is to coordinate the development of the new employee app, and provides a detailed explanation of the new tabs. After this, create a townhall titled Employee App Initiative Briefing, giving it a short description that explains the session will introduce the goals and direction of this new internal initiative. Schedule it for November 17, 2025, from 10:30 AM to 2:30 PM (UTC), make it available only to the organization, and add as co-organizers the members of the channel whose job titles are Senior Developer or UX Designer.\\

\textbf{Oracle Tools:} \texttt{list\_teams}, \texttt{list\_users}, \texttt{create\_channel}, \texttt{list\_teams\_apps}, \texttt{add\_tabs\_to\_channels}, \texttt{send\_channel\_message}, \texttt{create\_virtual\_event\_townhall}
\end{taskbox}

\subsection{Hybrid}
\label{app:cat_hybrid}

The \textbf{Hybrid} category represents the most complex class of tasks, requiring the agent to usually operate across two of the seven distinct domains simultaneously. This significantly increases the complexity of the environment, with an average of 40 tables compared to the mean of 25 across single-domain tasks. These scenarios simulate realistic enterprise workflows where actions in one system trigger requirements in another, demanding high-level planning and state tracking across disparate APIs.

For example, a task might require the agent to \textit{``Check if a product's warranty extends beyond 2025 in CSM; if so, log a customer interaction and immediately schedule a 'Warranty Discussion' on the Sales Calendar.''} This compels the agent to retrieve information from the CSM database (warranty status), make a decision based on that data, perform a write operation in CSM (logging the interaction), and then context-switch to the Calendar API to schedule an event using details derived from the CSM record. Success depends on maintaining state consistency across both platforms and correctly mapping entities (like Product IDs to Event Descriptions) between them.

\begin{taskbox}[Hybrid Example Task]
\textbf{System Prompt:} \\
You are an integrated automation agent for a hybrid environment managing both Google Calendar and Customer Service Management (CSM). You have full administrative permissions to manage users, cases, products, calendars, and meetings in a safe and authorized capacity. Do not ask for confirmation before taking action. \\

\textbf{User Prompt:} Please check the installed product with serial number P55-940931-6065 to confirm whether its warranty extends beyond 2025. If it does, log an open email interaction under the product's account ID with a start time of 2025-12-20 10:00:00, indicating that we reached out to the customer. Then, immediately schedule a `Warranty Discussion' event on my calendar for sales activities. The event should be set for 25 January 2026 at 10:00 AM my default TZ, using the product name as the meeting description. \\

\textbf{  Tools:} \texttt{find\_installed\_product\_by\_serial}, \texttt{register\_new\_interaction}, \texttt{get\_calendar\_list}, \texttt{find\_product\_by\_id}, \texttt{list\_settings}, \texttt{create\_event}
\end{taskbox}

\section{Rollout examples (concise)}
\label{app:errors}
To illustrate the nature of benchmark tasks and the types of failures models exhibit, we present abridged Claude-Sonnet-4.5 \citep{anthropic2025sonnet45} rollout examples with summary of the user task, system policy and tools, drawn from the ITSM, CSM and HR domains.

\begin{taskbox}[ITSM Example]

\subsection{\texorpdfstring{Case Study: ITSM --- ``Create KB and Link''
(\textit{claude-sonnet-4-5}, 0/2 verifiers passed)}%
{Case Study: ITSM --- "Create KB and Link" (claude-sonnet-4-5, 0/2 verifiers passed)}}%
\label{case-study-itsm-create-kb-and-link-claude-sonnet-4-5-02-verifiers-passed}

\subsection{Task (condensed)}\label{task-condensed}

Kenji Tanaka (agent, Acme Corp) resolved incident INC0000004 (VPN
connection failure) without referencing a knowledge article. He must
draft a new internal KB article titled ``VPN Connection Failure Guide''
and link it to the incident.
\textbf{Relevant policy (excerpts):}
\begin{itemize}
  \item \textbf{\S7 Knowledge Creation:}
    \textit{``\ldots the Agent must create and link a new
    \textbf{knowledge draft} before final closure.''}
  \item \textbf{\S1 General Constraint:}
    \textit{``Never assume or fabricate IDs \ldots\ rely solely on
    verified API results. The same is the case for optional or default
    arguments.''}
\end{itemize}
\subsection{Hidden Challenge: Duplicate Incident Numbers}%
\label{hidden-challenge-duplicate-incident-numbers}

Two incidents share external ID \texttt{INC0000004} across different
tenants:

\medskip
\begin{longtable}[l]{@{}
  >{\raggedright\arraybackslash\small}p{0.14\linewidth}
  >{\raggedright\arraybackslash\small}p{0.13\linewidth}
  >{\raggedright\arraybackslash\small}p{0.24\linewidth}
  >{\raggedright\arraybackslash\small}p{0.13\linewidth}
  >{\raggedright\arraybackslash\small}p{0.20\linewidth}@{}}
\toprule
\small Internal ID & \small Org & \small Description & \small Status & \small Assignee \\
\midrule
\endhead
\bottomrule
\endlastfoot
\texttt{INC\_004} & TechCorp  & Network connectivity issues &
  \textbf{new}      & Elena Petrov \\[4pt]
\texttt{INC\_011} & Acme Corp & VPN connection failure &
  \textbf{resolved} & \textbf{Kenji Tanaka} $\checkmark$ \\
\end{longtable}
\medskip

\texttt{find\_incident\_by\_number("INC0000004")} returns
\texttt{INC\_004} (first DB hit). Recovery requires a follow-up:\\
\texttt{list\_incidents(number="INC0000004",\ status="resolved",\
assigned\_to="USER\_009")} $\rightarrow$ \texttt{INC\_011}.

\subsection{Gold Trajectory}\label{gold-trajectory}

\begin{enumerate}
  \item \texttt{get\_user\_using\_name("Kenji",\ "Tanaka")}
        $\rightarrow$ \texttt{USER\_009}
  \item \texttt{find\_incident\_by\_number("INC0000004")}
        $\rightarrow$ \texttt{INC\_004}
        $\rightarrow$ \textbf{detect mismatch} (wrong status,
        description, assignee)
  \item \texttt{list\_incidents(number=...,\ status="resolved",\
        assigned\_to="USER\_009")} $\rightarrow$ \texttt{INC\_011}
        $\checkmark$
  \item \texttt{find\_incident\_knowledge\_links("INC\_011")}
        $\rightarrow$ no existing links
  \item \texttt{create\_knowledge\_article(...,\
        state=}\textbf{``draft''}\texttt{,\
        visibility="internal",\ owner\_id="USER\_009")}
  \item \texttt{link\_knowledge\_to\_incident("INC\_011",\
        "KB\_006",\ used\_as="resolution")}
\end{enumerate}

\subsection{Agent Behavior}\label{agent-behavior}

The agent called \texttt{find\_incident\_by\_number}, received
\texttt{INC\_004}, and accepted it without validation. It never called
\texttt{list\_incidents} to explore the difference. It then created the
KB article with \texttt{state="published"} (the tool default) and linked
it to \texttt{INC\_004}.

\subsection{Failure Analysis}\label{failure-analysis}

\textbf{Failure 1 --- Wrong incident.} \texttt{INC\_004} contradicted
the task on three observable signals: wrong status (\texttt{new}
vs.\ \textit{resolved}), wrong description, and wrong assignee. The
agent treated number-match as identity-confirmation and never
cross-validated. Verifier checks \texttt{incident\_id = "INC\_011"} in
the link table $\rightarrow$ \textbf{fail}.

\textbf{Failure 2 --- Wrong KB state.} The tool's default is
\texttt{state="published"}. Both \S7 and the user's verb
(\textit{``drafts''}) mandate \texttt{state="draft"}. \S1 explicitly
prohibits accepting defaults without policy verification. The agent
applied the default silently. Verifier checks \texttt{state = "draft"}
$\rightarrow$ \textbf{fail}.

\subsection{Summary}\label{summary}

\medskip
\begin{longtable}[l]{@{}
  >{\raggedright\arraybackslash}p{0.22\linewidth}
  >{\raggedright\arraybackslash}p{0.22\linewidth}
  >{\raggedright\arraybackslash}p{0.22\linewidth}
  >{\raggedright\arraybackslash}p{0.28\linewidth}@{}}
\toprule
                     & Expected                   & Agent              & Impact \\
\midrule
\endhead
\bottomrule
\endlastfoot
Incident ID          & \texttt{INC\_011}          & \texttt{INC\_004}  & KB link verifier fails \\[4pt]
KB state             & \texttt{draft}             & \texttt{published} & New KB verifier fails \\[4pt]
Disambiguation step  & \texttt{list\_incidents(...)} called & Never called & Root cause of wrong incident \\
\end{longtable}
\medskip

Both failures share the same pattern: \textbf{accepting the first
plausible result without cross-validating against task context or policy
constraints.}

\end{taskbox}
\begin{taskbox}[CSM Example]
\subsection[Case Study: CSM --- ``KB Remediation and Case Setup''
  (claude-sonnet-4-6, 4/5 verifiers passed)]{%
  Case Study: CSM --- ``KB Remediation and Case Setup''\\
  \normalfont\normalsize\emph{(claude-sonnet-4-6, 4/5 verifiers passed)}}
\label{case-study-csm-kb-remediation-and-case-setup}

\subsection{Task (condensed)}
\label{task-condensed}

An agent must link a relevant knowledge article to case CS-0000002 and
set up the assignee. The case involves a NetApp FAS2750 product issue.
Joanne Simpson will handle the case under a new ``Case Management''
support group.

\textbf{Relevant policy (excerpts):}
\begin{itemize}
  \item \textbf{KB Linking:} \emph{``Articles must be in state =
    \texttt{published} \ldots{} when the knowledge is found through
    automated search it should be linked as \texttt{suggested}.''}
  \item \textbf{Case State:} \emph{``Once case linked to a knowledge
    article marked the state = \texttt{pending}.''}
  \item \textbf{Assignment:} \emph{``\texttt{assigned\_to} must \ldots{}
    be member of \texttt{assignment\_group\_id}.''}
\end{itemize}

\subsection{Hidden Challenges}
\label{hidden-challenges}

Three compounding complexities are not stated in the user prompt and must
be inferred from system policy:

\begin{itemize}
  \item \textbf{KB state remediation:} KB-0000197 is \texttt{retired};
    must be updated to \texttt{published} before linking.
  \item \textbf{Group creation:} \texttt{"Case Management"} does not
    exist; must be created with \texttt{type="support"}.
  \item \textbf{Lifecycle transition:} KB linkage unconditionally requires
    \texttt{update\_case(state="pending")}.
\end{itemize}

\subsection{Gold Trajectory}
\label{gold-trajectory}

\begin{enumerate}
  \item \texttt{search\_cases(number="CS-0000002")}
        $\to$ \texttt{case\_id=2}, \texttt{product\_id=130}
  \item \texttt{retrieve\_knowledge(product\_id=130)}
        $\to$ \texttt{knowledge\_id=197}, \texttt{state="retired"}
  \item \texttt{update\_knowledge(knowledge\_id=197, state="published")}
        $\to$ KB now usable
  \item \texttt{link\_case\_knowledge(case\_id=2, knowledge\_id=197,
        used\_as="suggested")}
        $\to$ link created
  \item \texttt{find\_user(name="Joanne Simpson")}
        $\to$ \texttt{user\_id=4}, \texttt{role="manager"},
        \texttt{active=1}
  \item \texttt{find\_user\_group(name="Case Management")}
        $\to$ \texttt{\{\}} (absent)
  \item \texttt{add\_new\_user\_group(name="Case Management",
        type="support", active=true)}
        $\to$ \texttt{group\_id=81}
  \item \texttt{add\_new\_group\_member(group\_id=81, user\_id=4)}
        $\to$ membership created
  \item \texttt{update\_case(case\_id=2, assignment\_group\_id=81,
        assigned\_to=4,}
        \textbf{\texttt{state="pending"}}\texttt{)}
        $\to$ case closed
\end{enumerate}

\subsection{Agent Behavior}
\label{agent-behavior}

The agent executed a near-perfect 5-turn trajectory. In Turn~1 it issued
three parallel lookups (case, user, group). In Turns~2--3 it correctly
pivoted from a text-based KB search (which returned wrong product
variants) to a \texttt{product\_id=130} filter, finding the retired
KB-0000197. In Turn~4 it published the KB and created the group in
parallel. In Turn~5 it added Joanne to the group, linked the KB article,
and updated the case assignment --- but omitted \texttt{state="pending"}
from the \texttt{update\_case} call.

\subsection{Failure Analysis}
\label{failure-analysis}

\textbf{Single failure --- missing state transition.} The agent's
\texttt{update\_case} call set \texttt{case\_id},
\texttt{assignment\_group\_id}, and \texttt{assigned\_to} correctly, but
did not include \texttt{state="pending"}. The case remained in
\texttt{state="new"}. The policy rule is explicit and unconditional: any
KB linkage event requires a transition to \texttt{pending}. The agent's
Turn~5 reasoning focused on ``update the case assignment'' without
revisiting the lifecycle rules --- a classic lifecycle-truncation failure
(Pattern~\#4). The \texttt{state} parameter and the \texttt{pending} enum
value are both present in the tool schema; no tool error or ambiguity
blocked the correct call.

\subsection{Summary}
\label{summary}

\begin{center}
\begin{tabular}{llll}
\toprule
\textbf{Check}           & \textbf{Expected}                         & \textbf{Agent}               & \textbf{Impact} \\
\midrule
\texttt{update\_case.state}   & \texttt{"pending"}               & omitted (\texttt{"new"})     & V5 fail \\
KB remediation                & \texttt{update\_knowledge(state="published")} & correct             & V1 pass \\
KB linkage                    & \texttt{used\_as="suggested"}    & correct                      & V2 pass \\
Group creation                & \texttt{type="support"}          & correct                      & V3 pass \\
Membership check              & \texttt{add\_new\_group\_member} before assignment & correct    & V4 pass \\
\bottomrule
\end{tabular}
\end{center}

\medskip

The failure isolates to a single omitted parameter on an otherwise
correct trajectory: \textbf{the agent completed the assignment but did
not apply the KB-linkage-triggered lifecycle rule.}
\end{taskbox}
\begin{taskbox}[HR Example]
\subsection{\texorpdfstring{Case Study: HR --- ``Wrap Up James Hill's Portal Access Case''
(\textit{claude-sonnet-4-5}, 1/3 verifiers passed)}%
{Case Study: HR --- "Wrap Up James Hill's Portal Access Case" (claude-sonnet-4-5, 1/3 verifiers passed)}}%
\label{case-study-hr-wrap-up-james-hills-portal-access-case-claude-sonnet-4-5-13-verifiers-passed}

\subsection{Task (condensed)}
\label{task-condensed}

Karen Watkins (admin) is told that James Hill's `Access issue with HR
portal account' has been resolved. She must wrap up his case and add a
follow-up technical issue survey using the first task to gather his
feedback.

\medskip

\textbf{Relevant policy (condensed):}
\begin{itemize}
  \item \textbf{§3.2 Closure Constraint:} \emph{``A case cannot move to a
    closed status until all mandatory tasks are inactive
    (\texttt{active=false}).''}
  \item \textbf{§3.3 Lifecycle:} \emph{``Valid transition:
    \texttt{awaiting\_approval} $\to$ \texttt{closed\_complete} (if
    approved).''}
  \item \textbf{§3.4 Approvals:} \emph{``The approval record
    \texttt{request\_status} transitions: \texttt{requested} $\to$
    \texttt{approved} / \texttt{rejected}.''}
  \item \textbf{§1 General Constraint:} \emph{``Do not ask for any
    information or confirmation from the user. Never assume or fabricate
    IDs.''}
\end{itemize}

\subsection{Hidden Challenge: Two Simultaneous Closure Prerequisites}
\label{hidden-challenge-two-simultaneous-closure-prerequisites}

``Wrap up'' maps to three ordered system operations --- none stated
literally in the prompt:

\begin{center}
\begin{tabular}{lll}
\toprule
\textbf{Step} & \textbf{Action} & \textbf{Policy Source} \\
\midrule
1 & Deactivate all active tasks (\texttt{active=false})
  & §3.2 --- prerequisite to closure \\
2 & Set \texttt{status='closed\_complete'}
  & §3.3 --- valid closure status for resolved cases \\
3 & Set \texttt{request\_status='approved'}
  & §3.4 --- clears the pending approval gate \\
\bottomrule
\end{tabular}
\end{center}

\medskip

The case seed state has \texttt{status='awaiting\_approval'} and
\texttt{request\_status='requested'} with two active tasks (ids 6 and
7). Steps 1--3 must all complete; failing any leaves the verifier
returning COUNT=0.

\subsection{Gold Trajectory}
\label{gold-trajectory}

\begin{enumerate}
  \item \texttt{get\_user\_using\_name("James",\ "Hill")}
        $\to$ \texttt{user\_id=8}
  \item \texttt{list\_hr\_cases(opened\_for="James Hill")}
        $\to$ \texttt{hr\_case\_id=3},
        \texttt{status='awaiting\_approval'},
        \texttt{request\_status='requested'}
  \item \texttt{list\_hr\_case\_tasks(parent\_case="3")}
        $\to$ Tasks 6 (\texttt{url}, active=True) and 7
        (\texttt{checklist}, active=True)
  \item \texttt{list\_surveys(question\_1="technical issue")}
        $\to$ \texttt{survey\_id=4}
  \item \texttt{update\_hr\_case\_task(hr\_case\_task\_id="6",\ active=false)}
        $\to$ Task 6 deactivated
  \item \texttt{update\_hr\_case\_task(hr\_case\_task\_id="7",\ active=false)}
        $\to$ Task 7 deactivated
  \item \texttt{update\_hr\_case(hr\_case\_id="3",
        status="closed\_complete",\ request\_status="approved")}
        $\to$ Case closed and approved
  \item \texttt{create\_survey\_instance(survey\_id=4,\ case\_task\_id=6,\ assigned\_to=8)}
        $\to$ Survey instance created \checkmark
\end{enumerate}

\subsection{Agent Behavior}
\label{agent-behavior}

The agent completed all four lookup steps correctly and created the
survey instance with the right parameters (V3 passes). It then called
\texttt{update\_hr\_case\_task} on task 6 --- but passed
\texttt{task\_type="survey"} and a new \texttt{short\_description}
instead of \texttt{active=false}. It never called
\texttt{update\_hr\_case}. The agent declared completion after six
turns, summarising the survey as something James Hill would complete in
the future.

\subsection{Failure Analysis}
\label{failure-analysis}

\textbf{Failure 1 --- Wrong parameters on \texttt{update\_hr\_case\_task}.}
The agent read ``add the appropriate follow-up technical issue survey
\textit{using the first task}'' as a directive to convert task~6 into a
survey-type task. It therefore called
\texttt{update\_hr\_case\_task(task\_type="survey",\ short\_description=\ldots{})}
rather than \texttt{update\_hr\_case\_task(active=false)}. The §3.2
Closure Constraint --- visible in the system prompt and signalled by
\texttt{update\_hr\_case\_task}'s presence in the tool set --- requires
deactivation, not type conversion. Task~6 remained \texttt{active=true};
the verifier checks \texttt{active=false} $\to$ \textbf{fail}.

\medskip

\textbf{Failure 2 --- \texttt{update\_hr\_case} never called.} The agent
reframed ``wrap up his case'' as a future activity for James Hill
(completing the survey) rather than an immediate system closure. Its
final summary reads: ``James can complete the survey as part of the case
wrap-up process.'' The agent stopped at survey creation and declared
success. \texttt{update\_hr\_case} was present in the tool set (a
planning signal), §3.3 specifies
\texttt{awaiting\_approval} $\to$ \texttt{closed\_complete} as the valid
transition, and \texttt{request\_status='requested'} was visible in the
\texttt{list\_hr\_cases} response. The case remained in
\texttt{awaiting\_approval}; the verifier checks
\texttt{status='closed\_complete'\ AND\ request\_status='approved'}
$\to$ \textbf{fail}.

\subsection{Summary}
\label{summary}

\begin{center}
\begin{tabular}{llll}
\toprule
\textbf{Check} & \textbf{Expected} & \textbf{Agent} & \textbf{Impact} \\
\midrule
Task 6 \texttt{active}        & \texttt{false}            & wrong params passed  & V1 fails \\
Case \texttt{status}          & \texttt{closed\_complete} & tool never called    & V2 fails \\
Case \texttt{request\_status} & \texttt{approved}         & tool never called    & V2 fails \\
Survey instance               & correct params            & correct              & V3 passes \\
\bottomrule
\end{tabular}
\end{center}

\medskip

Both failures stem from the same root: \textbf{natural-language business
verbs (``wrap up'', ``using the first task'') misread as content
directives rather than lifecycle commands}, causing the agent to act on
a plausible surface reading while ignoring the policy-defined closure
sequence.
\end{taskbox}

\section{Additional Analysis and Results}
\label{app:additional}
\begin{figure*}[h]
 \centering
 \includegraphics[width=1.0\linewidth]{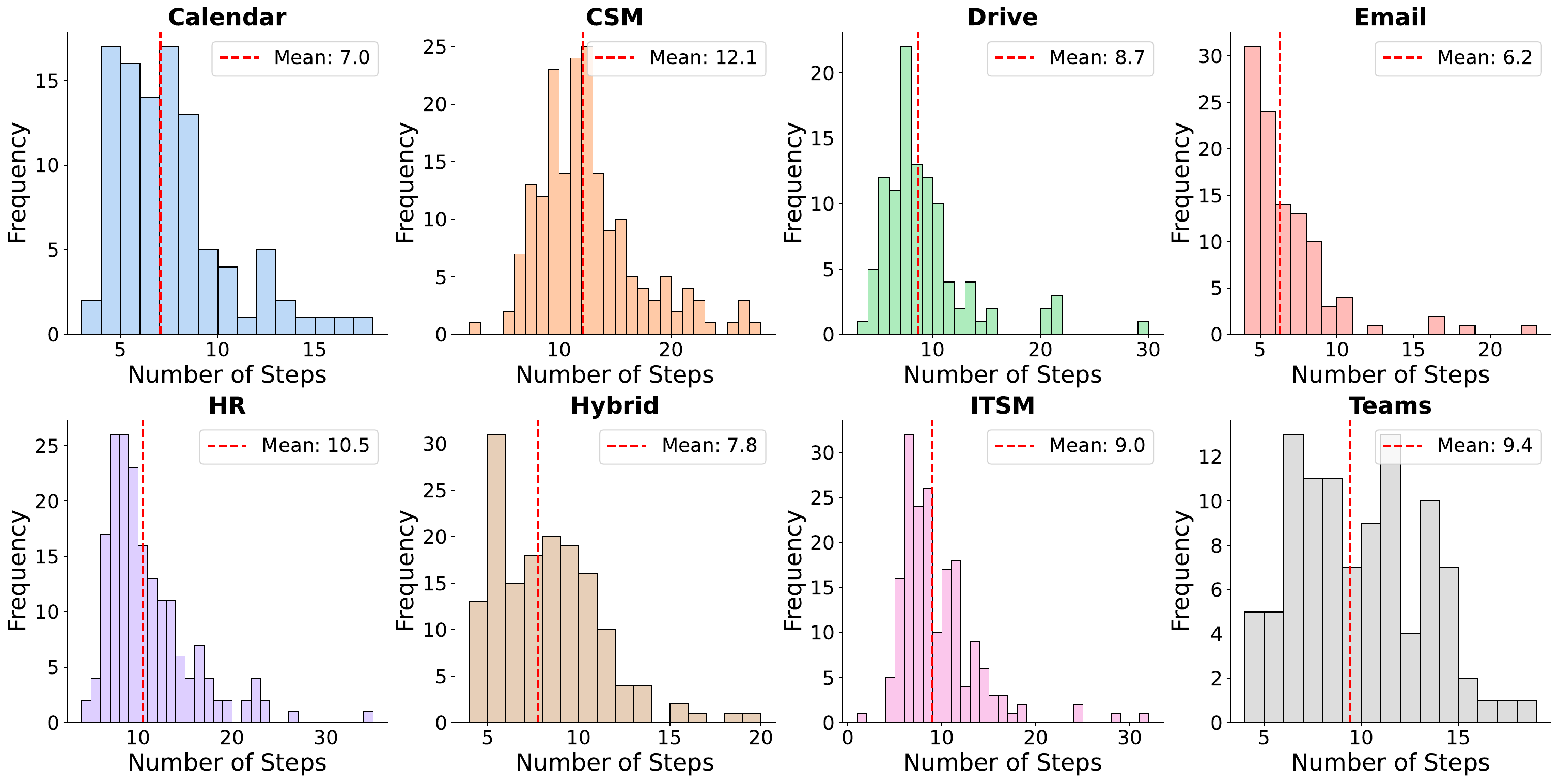}
 \caption{\textbf{Distribution of task complexity across domains in \dataset{}.}
Histograms show the distribution for number of tool-execution steps required to complete tasks in each domain. Red dashed lines indicate the mean steps per domain. The results highlight substantial variation in interaction depth, with CSM and HR exhibiting longer, heavier-tailed execution traces, while Email and Calendar tasks tend to require shorter, more tightly distributed action sequences. Hybrid and ITSM workflows show moderate to high variance.}
 \label{fig:tasks_len_distribution}
\end{figure*}

We present additional analysis on task complexity in \Cref{fig:tasks_len_distribution}, \Cref{tab:verifier_analysis,tab:main_results_appendix}, and full results in \Cref{tab:main_results}.

\begin{table*}[h!]
\centering
\resizebox{\textwidth}{!}{%
\begin{tabular}{lcccccccccccccccccc}
\toprule
\multirow{2}{*}{\textbf{Model}} & \multicolumn{2}{c}{\textbf{Teams}} & \multicolumn{2}{c}{\textbf{CSM}} & \multicolumn{2}{c}{\textbf{Email}} & \multicolumn{2}{c}{\textbf{ITSM}} & \multicolumn{2}{c}{\textbf{Calendar}} & \multicolumn{2}{c}{\textbf{HR}} & \multicolumn{2}{c}{\textbf{Drive}} & \multicolumn{2}{c}{\textbf{Hybrid}} & \multicolumn{2}{c}{\textbf{Overall}} \\
\cmidrule(lr){2-3} \cmidrule(lr){4-5} \cmidrule(lr){6-7} \cmidrule(lr){8-9} \cmidrule(lr){10-11} \cmidrule(lr){12-13} \cmidrule(lr){14-15} \cmidrule(lr){16-17} \cmidrule(lr){18-19}
& \textbf{O} & \textbf{VL} & \textbf{O} & \textbf{VL} & \textbf{O} & \textbf{VL} & \textbf{O} & \textbf{VL} & \textbf{O} & \textbf{VL} & \textbf{O} & \textbf{VL} & \textbf{O} & \textbf{VL} & \textbf{O} & \textbf{VL} & \textbf{O} & \textbf{VL} \\
\midrule
\rowcolor{gray!15} \multicolumn{19}{c}{\textit{Closed Source Models}} \\
Claude Opus 4.5 \citep{anthropic2025sonnet45} & 50.0 & 84.2 & 34.2 & 66.9 & 51.9 & 72.6 & 23.9 & 54.8 & 43.3 & 73.0 & 32.1 & 67.0 & 49.5 & 75.3 & 30.7 & 61.7 & 37.0 & 67.7 \\
Gemini-3-Flash \citep{comanici2025gemini25pushingfrontier} & 47.3 & 81.0 & 35.0 & 68.1 & 44.3 & 70.0 & 28.5 & 58.3 & 30.5 & 67.0 & 12.6 & 54.9 & 49.7 & 76.7 & 24.2 & 58.3 & 31.4 & 65.0 \\
GPT-5.2 (High) \citep{openai2025gpt5} & 31.0 & 66.3 & 34.8 & 67.6 & 51.0 & 72.6 & 21.7 & 49.9 & 38.5 & 71.2 & 25.0 & 60.1 & 40.0 & 67.9 & 22.2 & 54.6 & 31.3 & 62.5 \\
Claude Sonnet 4.5 \citep{anthropic2025sonnet45} & 51.0 & 81.9 & 16.7 & 55.7 & 51.3 & 71.7 & 17.6 & 49.9 & 34.6 & 70.9 & 21.6 & 61.4 & 52.1 & 76.0 & 28.1 & 61.4 & 34.1 & 66.1 \\
GPT-5 \citep{openai2025gpt5} & 26.3 & 59.7 & 36.4 & 62.6 & 49.0 & 71.2 & 18.9 & 38.4 & 41.3 & 71.5 & 17.9 & 53.9 & 34.0 & 62.3 & 23.5 & 53.3 & 30.9 & 59.1 \\
Gemini-3-Pro \citep{comanici2025gemini25pushingfrontier} & 43.0 & 77.1 & 27.7 & 64.6 & 33.7 & 63.2 & 22.2 & 56.3 & 28.8 & 66.1 & 12.5 & 53.1 & 46.7 & 73.0 & 22.9 & 56.6 & 27.5 & 62.2 \\
GPT-5.2 (Low) \citep{openai2025gpt5} & 25.0 & 55.9 & 21.2 & 53.6 & 43.3 & 68.4 & 6.7 & 20.9 & 28.8 & 63.4 & 13.0 & 37.5 & 26.7 & 58.3 & 20.9 & 48.1 & 21.1 & 47.8 \\
GPT-5-Mini \citep{openai2025gpt5} & 25.7 & 55.2 & 15.8 & 51.6 & 47.4 & 71.1 & 8.9 & 29.7 & 28.8 & 65.3 & 10.7 & 42.7 & 23.8 & 56.3 & 22.5 & 51.4 & 22.9 & 52.9\\
Gemini-2.5-Pro \citep{comanici2025gemini25pushingfrontier} & 39.3 & 66.8 & 11.6 & 48.0 & 31.1 & 59.8 & 13.9 & 44.8 & 12.5 & 52.2 & 4.9 & 44.9 & 27.0 & 57.9 & 19.6 & 53.2 & 20.0 & 53.5 \\
\rowcolor{gray!15} \multicolumn{19}{c}{\textit{Open Source Models}} \\
DeepSeek-V3.2 (High)~\citep{deepseekai2024deepseekv3technicalreport} & 37.0 & 67.3 & 14.1 & 51.7 & 47.1 & 67.3 & 16.1 & 47.3 & 21.2 & 55.8 & 16.3 & 50.1 & 35.2 & 59.3 & 22.9 & 52.1 & 23.8 & 54.7 \\
GPT-OSS-120B (High)~\citep{openai2025gptoss120bgptoss20bmodel} & 32.0 & 61.8 & 16.3 & 52.8 & 42.3 & 64.6 & 6.1 & 24.1 & 35.6 & 66.8 & 16.3 & 49.3 & 41.0 & 67.4 & 19.6 & 52.4 & 23.1 & 52.1 \\
DeepSeek-V3.2 (Medium)~\citep{deepseekai2024deepseekv3technicalreport} & 35.7 & 62.6 & 15.4 & 44.4 & 45.8 & 66.0 & 9.6 & 27.9 & 21.5 & 49.0 & 15.0 & 48.0 & 27.6 & 55.3 & 22.9 & 55.9 & 24.2 & 51.1 \\
Kimi-K2-Thinking~\citep{kimiteam2025kimik2openagentic} & 30.0 & 71.2 & 7.1 & 40.6 & 51.0 & 69.6 & 12.2 & 35.5 & 15.4 & 54.7 & 8.2 & 39.5 & 39.6 & 61.4 & 15.7 & 49.5 & 22.4 & 52.8 \\
Qwen3-235B (Inst.)~\citep{qwen3} & 28.0 & 58.3 & 4.7 & 30.3 & 38.1 & 64.0 & 9.3 & 35.8 & 15.7 & 54.1 & 7.8 & 39.1 & 23.8 & 55.4 & 17.7 & 47.6 & 18.1 & 48.1 \\
Qwen3-30B (Think)~\citep{qwen3} & 22.0 & 52.4 & 5.4 & 37.9 & 51.9 & 72.0 & 6.7 & 35.3 & 18.3 & 61.7 & 7.6 & 38.5 & 25.7 & 51.7 & 15.7 & 48.2 & 19.1 & 49.7\\
Qwen3-4B (Think)~\citep{qwen3} & 24.0 & 53.1 & 3.8 & 32.8 & 38.4 & 63.1 & 5.6 & 32.3 & 5.8 & 48.4 & 7.1 & 41.4 & 21.9 & 55.3 & 15.8 & 47.6 & 15.3 & 46.8 \\
\bottomrule
\end{tabular}%
}
\caption{\textbf{Overall task completion performance on \ours{}.} (Oracle Mode) We report the percentage of tasks successfully completed by each model in oracle tool mode, broken down by domain. A task is considered successful only if all outcome verification checks pass. Results highlight substantial performance degradation on long-horizon and cross-domain (Hybrid) workflows, indicating persistent improvement gap. VL: Verifier Level, O: Overall.}
\label{tab:main_results_appendix}
\end{table*}
\begin{table}[htbp]
\centering
\begin{tabular}{l|rr|rr|rr}
\hline
\multirow{2}{*}{\textbf{Domain}} & \multicolumn{2}{c|}{\textbf{Task Completion}} & \multicolumn{2}{c|}{\textbf{Integrity Constraints}} & \multicolumn{2}{c}{\textbf{Policy Compliance}} \\
& \textbf{Samples} & \textbf{Score} & \textbf{Samples} & \textbf{Score} & \textbf{Samples} & \textbf{Score} \\
\hline
Teams    & 100 & 82.3 & 20 & 90.0 & 8   & 81.3 \\
CSM      & 183 & 58.3 & 30 & 70.0 & 120 & 45.5 \\
Email    & 103 & 70.9 & 17 & 76.5 & 12  & 79.2 \\
ITSM     & 176 & 53.7 & 30 & 50.6 & 70  & 30.0 \\
Calendar & 104 & 71.5 & 8  & 62.5 & 19  & 76.1 \\
HR       & 182 & 63.2 & 47 & 56.0 & 94  & 50.4 \\
Drive    & 105 & 75.1 & 26 & 87.8 & 6   & 66.7 \\
Hybrid   & 152 & 60.0 & 23 & 72.5 & 18  & 61.1 \\
\hline
\end{tabular}
\caption{Verifier pass rates across domains by category. Models achieve highest scores on Integrity Constraints (system state validity), followed by Task Completion, with Policy Compliance showing the lowest performance. Sample counts vary across categories, with heavier domains such as CSM, ITSM and HR with more emphasis on Integrity Constraints and Policy compliance.}
\label{tab:verifier_analysis}
\end{table}

\section{Impact Statement} \label{app:impact}
This work contributes a benchmark and evaluation framework that targets a core challenge in modern AI systems: reliable planning and execution over real tools with persistent state. As LLM-based agents are increasingly deployed to automate workflows involving scheduling, communication, document management, and operational support, understanding their failure modes under realistic constraints is critical.

\textbf{Positive Impacts:}
\ours{} provides the research community with a rigorous, reproducible testbed for studying agentic planning, tool selection, and error recovery. By emphasizing outcome-based verification and safety-critical constraints, the benchmark encourages the development of agents that are not only capable, but also reliable and auditable. Progress enabled by this benchmark may lead to safer automation of repetitive and high-friction workflows, reducing human burden and enabling practitioners to focus on higher-level decision-making. More broadly, \ours{} advances the study of general-purpose tool use and planning, independent of any single enterprise platform.

\textbf{Potential Risks and Limitations:}
As with any benchmark that evaluates execution over tool interfaces, there is a risk that systems trained exclusively to optimize benchmark performance may overfit to specific task patterns or verification criteria. Additionally, increased automation of operational workflows may raise concerns around over-reliance on AI systems, particularly if failures are not well understood or monitored. Finally, training and evaluating large models for agentic behavior incurs computational cost, which has environmental implications that should be considered alongside performance gains.

We mitigate these risks by releasing \ours{} as a diagnostic benchmark, emphasizing analysis over leaderboard ranking, and by ensuring that all environments are synthetic and sandboxed, with no access to proprietary or sensitive data. We hope this work supports responsible research into agentic systems and helps the community build agents that are not only more capable, but also more dependable.

 
\end{document}